\documentclass[letterpaper]{article} 
\usepackage{aaai2026}  
\usepackage{times}  
\usepackage{helvet}  
\usepackage{courier}  
\usepackage[hyphens]{url}  
\usepackage{graphicx} 
\usepackage{natbib}  
\usepackage{caption} 
\usepackage{algorithm}
\usepackage{algorithmic}
\usepackage{amsmath}
\usepackage{amssymb}
\usepackage{multirow}
\usepackage{makecell}
\usepackage{xcolor}

\newcommand{\fullwidthtitle}[1]{
  \twocolumn[{
    \centering
    \vspace*{2\baselineskip} 
    {\LARGE\bfseries #1\par} 
    \vspace{\baselineskip} 
  }]
}

\usepackage{newfloat}
\usepackage{listings}
\DeclareCaptionStyle{ruled}{labelfont=normalfont,labelsep=colon,strut=off} 
\floatstyle{ruled}
\newfloat{listing}{tb}{lst}{}
\floatname{listing}{Listing}
\title{CloudMamba: Grouped Selective State Spaces for Point Cloud Analysis}
\author{
    Kanglin Qu\textsuperscript{\rm 1},
    Pan Gao\textsuperscript{\rm 1}$^*$,
    Qun Dai\textsuperscript{\rm 1}\thanks{Corresponding author.}
    Zhanzhi Ye\textsuperscript{\rm 1},
    Rui Ye\textsuperscript{\rm 2},
    Yuanhao Sun\textsuperscript{\rm 3}
}
\affiliations{
    \textsuperscript{\rm 1}College of Artifcial Intelligence, Nanjing University of Aeronautics and Astronautics, Nanjing, 211106, China\\
    \textsuperscript{\rm 2}College of Artificial Intelligence, Nanjing Agricultural University, Nanjing, 210095, China\\
    \textsuperscript{\rm 3}School of Mathematical Sciences, Beijing University of Posts and Telecommunications, Beijing 100876, China\\
    klinqu@163.com, \{Pan.Gao, daiqun, yezhanzhi\}@nuaa.edu.cn, yerui@njau.edu.cn, sunyh@bupt.edu.cn
}

\usepackage{bibentry}

\begin{document}

\maketitle

\begin{abstract}
Due to the long-range modeling ability and linear complexity property, Mamba has attracted considerable attention in point cloud analysis. Despite some interesting progress, related work still suffers from imperfect point cloud serialization, insufficient high-level geometric perception, and overfitting of the selective state space model (S6) at the core of Mamba. To this end, we resort to an SSM-based point cloud network termed CloudMamba to address the above challenges. Specifically, we propose sequence expanding and sequence merging, where the former serializes points along each axis separately and the latter serves to fuse the corresponding higher-order features causally inferred from different sequences, enabling unordered point sets to adapt more stably to the causal nature of Mamba without parameters. Meanwhile, we design chainedMamba that chains the forward and backward processes in the parallel bidirectional Mamba, capturing high-level geometric information during scanning. In addition, we propose a grouped selective state space model (GS6) via parameter sharing on S6, alleviating the overfitting problem caused by the computational mode in S6. Experiments on various point cloud tasks validate CloudMamba's ability to achieve state-of-the-art results with significantly less complexity.
\end{abstract}

\begin{links}
    \link{Code}{https://github.com/Point-Cloud-Learning/CloudMamba}
    \link{Proceedings version}{https://ojs.aaai.org/index.php/AAAI/article/view/*****}
\end{links}

\section{Introduction}
\label{Section1}

The attention mechanism \cite{1,2} is gradually replacing the convolution as a dominant operator in point cloud analysis with its ability to achieve global modeling, but its quadratic complexity causes unbearable computational overheads. To handle this challenge, existing attention-based networks \cite{6,7,9,11} perform attention interactions in local neighborhoods or within windows, which, however, impose constraints on receptive fields. 
Recently, state space models (SSMs) \cite{13,15,16,18} have shown great potential in natural language processing (NLP).  Mamba \cite{19} achieves flexible selection of relevant information in a data-dependent manner by the selective state space model (S6), further enhancing long-range modeling capability. Moreover, Mamba with linear complexity adopts a hardware-aware algorithm inspired by FlashAttention \cite{20}, significantly improving training and inference efficiencies. 

\begin{table}[t]
  \centering
  \small
\setlength{\tabcolsep}{3.5mm}{
  \begin{tabular}{ccc}
    \hline
Serialization & Grid size & OA (\%)  \\
    \hline
Hilbert + Trans-Hilbert & 0.010 & 90.84 \\
Hilbert + Trans-Hilbert & 0.015 & 92.82  \\
Hilbert + Trans-Hilbert & 0.020 & 89.13  \\
Sequence expanding \& merging & / & \textbf{93.65} \\
    \hline
  \end{tabular}}
  \caption{Experimental results of different serialization methods in our network on ModelNet40 dataset.}
  \label{Tab0}
\end{table}

\begin{figure}[t]
\centering
\includegraphics[width=3.3in]{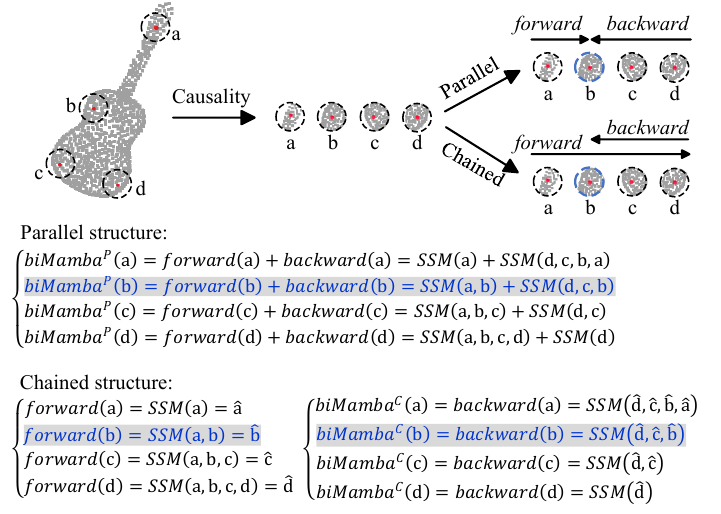}
\caption{Inference process of the bidirectional Mamba with different structures, where the blue equations denote the inference processes of both bidirectional Mamba for the point \textit{b}, respectively. In the backward inference of the chained structure, the previous points perceive the high-level structural semantics that are inferred from the forward Mamba.}
\label{Fig0}
\end{figure}
\begin{figure}[t]
\centering
\includegraphics[width=3.3in]{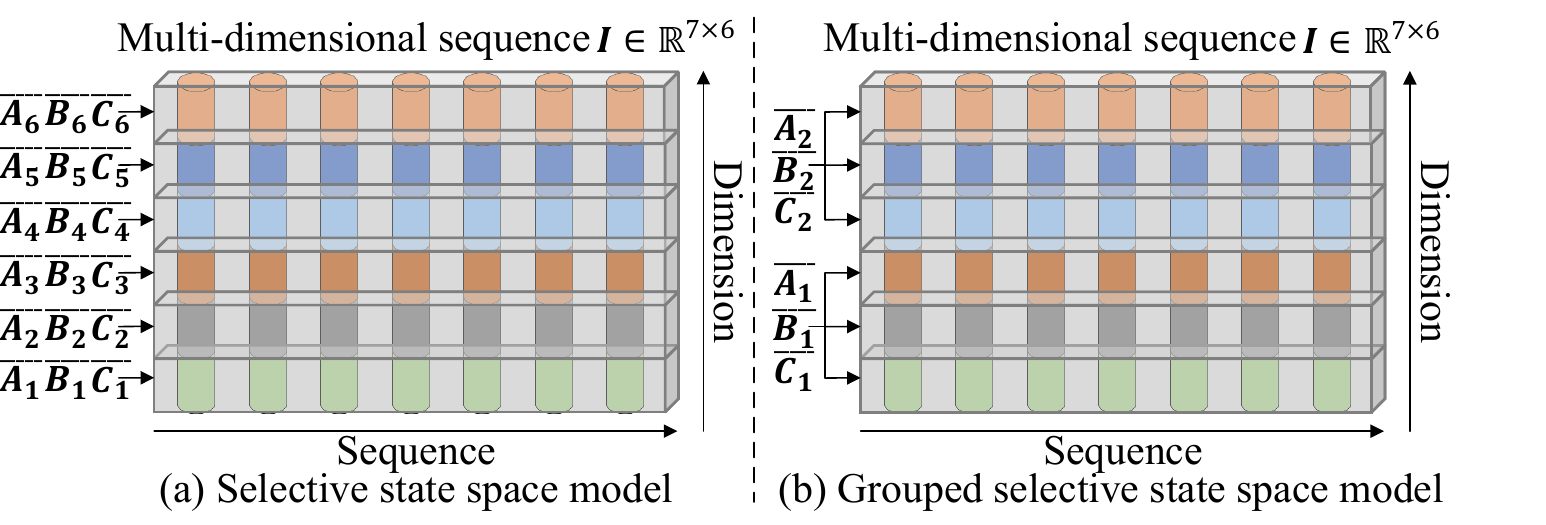}
\caption{Computational modes of S6 and GS6, where GS6's grouping rate is 3. A multi-dimensional sequence $\boldsymbol{I} \in {\mathbb{R}^{7 \times 6}}$ is used as an example, with the subscript denoting the parameter to be used for a dimension or a set of dimensions.}
\label{Fig1}
\end{figure}

Given Mamba's success in NLP, some works \cite{37,38,39,40} attempt to transplant this success from language modeling to point cloud analysis. In this process, despite some significant progress, these efforts still struggle to achieve satisfactory results due to the following three aspects:
\begin{itemize}
\item \textbf{Point cloud serialization.} It is crucial to build inter-point structure dependencies in a point sequence by serialization for Mamba's causal inference, and existing strategies are mostly based on space-filling curves. However, the reliability of the structural dependencies built by space-filling curves is highly sensitive to the setting of the grid size, as shown in Tab. \ref{Tab0}. Besides, as a pioneering study based on Mamba, PointMamba \cite{38} concatenates the serialization results of the Hilbert curve and its variant Trans-Hilbert \cite{29}. However, this practice has the following shortcomings: (1) a longer sequence induced by the concatenation introduces redundancy and negatively affects efficiency, and (2) concatenating serialized sequences with different spatial relationships tends to cause confusion. Hence, existing efforts are imperfect in point cloud serialization.
\item \textbf{High-level geometric perception.} While Mamba achieves superior long-range modeling via the selection mechanism, it is only unidirectional and not applicable to visual data requiring global learning. Therefore, some works \cite{39,40} introduce the bidirectional Mamba with a parallel structure from Vision Mamba \cite{31}. However, this structure limits the expressivity of networks. As illustrated in Fig. \ref{Fig0}, the parallel bidirectional Mamba provides each point with a global receptive field through the forward and backward inference, \textit{e.g.}, the point \textit{b} can interact with \textit{a}, \textit{c}, and \textit{d}. Nevertheless, we argue that this type of inference on primitive low-order geometric features can result in insufficient high-level geometric perception.
\item \textbf{Computational mode in S6.} Existing Mamba-based point cloud works \cite{38,39,40} take S6 as the core component, where each dimension is learned by a separate set of parameters when dealing with multi-dimensional sequences, as shown in Fig. \ref{Fig1}(a). This computational mode results in overfitting in causal inference due to excessive parameters in each dimension.
\end{itemize}

According to the above analyses, we propose a novel SSM-based point cloud network termed CloudMamba, which obeys the following designs to address the above challenges encountered in related works in order to greatly promote the development of Mamba in the point cloud domain:

\begin{itemize}
\item \textbf{Sequence expanding \& merging.} We propose sequence expanding, which serializes points along each axis separately, and sequence merging, which fuses the corresponding higher-order features causally inferred from different sequences. The sequence expanding directly builds structural dependencies in multiple perspectives from different axes, which not only overcomes the parameter sensitivity in space-filling curves, but also avoids the inefficiency and causal confusion caused by concatenating different serialization results. Furthermore, it captures rich geometric information by integrating with the sequence merging. Together, these enable unordered point sets to adapt more stably to the causal nature of Mamba without parameters, as shown in Tab. \ref{Tab0}.

\item \textbf{ChainedMamba.} We propose a chained bidirectional Mamba termed chainedMamba, which chains the forward and backward processes in the parallel bidirectional Mamba. In this case, due to the chaining property, the point \textit{b} could interact with the high-level structural semantics of $d$, $c$, $b$,  \textit{i.e.}, $\hat d$, $\hat c$, and $\hat b$ as shown in Fig. \ref{Fig0}. By utilizing these high-level structural semantics inferred from low-order geometric features in the forward Mamba during backward inference, it can achieve superior high-level geometric perception, as demonstrated in Tab. \ref{Tab6}.
\item \textbf{GS6.} We propose a grouped selective state space model (GS6) by sharing a same set of parameters across several dimensions, which alleviates the overfitting caused by the computational mode in S6, as illustrated in Fig. \ref{Fig1}(b).
\end{itemize}

Finally, we conduct experiments on point cloud recognition ModelNet40 \cite{22} and ScanObjectNN \cite{23}, part segmentation ShapeNet \cite{24}, as well as  semantic segmentation S3DIS \cite{63}. Experimental comparisons show CloudMamba is able to obtain state-of-the-art accuracy with linear complexity.



In summary, the contributions of this paper are fourfold:

\noindent
(1) A novel SSM-based point cloud network, CloudMamba, is constructed upon Mamba, which achieves state-of-the-art results in various point cloud tasks with linear complexity.

\noindent
(2) A serialization method consisting of sequence expanding and sequence merging is proposed, which builds structural dependencies directly from the coordinate axes defining spatial locations and captures rich geometric information by merging the sequences generated along different axes, making unordered point sets better adapted to the causal nature of Mamba without parameters.

\noindent
(3) A chained bidirectional Mamba termed chainedMamba is designed by chaining the forward and backward processes in the parallel bidirectional Mamba. This simple yet effective refinement achieves excellent high-level geometric perception.

\noindent
(4) A variant GS6 on S6 is proposed by adopting the parameter sharing, which mitigates overfitting in S6. To our best knowledge, all existing Mamba-based methods are based on S6, and GS6 is the first exploration of optimizing the computational mode in S6.

\section{Related work}
\label{Section2}

\subsection{SSMs and Mamba}
\label{Section2.2}

SSMs are mathematical models used in control theory to describe dynamic systems, which have recently been introduced to deep learning and have been developed considerably in long sequence modeling. Initially, Gu \textit{et al.} \cite{13} designed a linear state space layer (LSSL) based on linear continuous-time state space representations and demonstrated its potential to deal with long-range dependencies in sequential data by combining it with the HiPPO initialization \cite{14}. However, LSSL requires excessive computational resources and cannot be used as a generic sequential modeling method. Hence, the structured state space sequence model (S4) \cite{15} normalizes the parameters into a diagonal structure to solve the computational bottleneck. Subsequently, many efforts have been made to improve SSMs, such as GSS \cite{16}, S5 \cite{17}, and DSS \cite{18}, among which S6 is a landmark work, featuring a selective scanning mechanism and linear complexity. To enable flexible integration into neural networks, S6 is combined with the simplified H3 architecture \cite{19} to build Mamba, which achieves impressive results in NLP. Later, several works extend Mamba to other research directions, including image recognition \cite{30,74}, medical image segmentation \cite{33,35}, and graph sequence modeling \cite{36}.

Recently, Mamba has also begun to be applied to point cloud analysis. Mamba3D \cite{37} focuses on enhancing local features using Mamba, but it does not take into consideration the unidirectional modeling property of Mamba as our network does. PointMamba \cite{38} and OctMamba \cite{39} employ space-filling curves to form causal sequences for Mamba. Different from them, our network directly adopts coordinate axes to build structural dependencies in multiple perspectives without parameters, overcoming the parameter sensitivity in space-filling curves. PCM \cite{40} employs the parallel bidirectional Mamba, resulting in insufficient high-level geometric perception. In our work, chainedMamba achieves excellent high-level geometric perception by chaining the forward and backward processes in the parallel bidirectional Mamba. Finally, all the previous works only apply directly Mamba to point cloud analysis, while we improve its S6 to mitigate potential overfitting.

\section{Preliminaries}
\label{Section3}

\subsection{SSMs}
\label{Section3.1}

SSMs are cyclic processes with latent states, which map a 1-D equation or sequence $x\left( t \right) \in {{\rm{\mathbb{R}}}^N}$ to $y\left( t \right) \in {{\rm{\mathbb{R}}}^N}$ by a latent state $h\left( t \right) \in {{\rm{\mathbb{R}}}^N}$. The process is mathematically denoted as a linear ordinary differential equation as follows
\begin{equation}
y\left( t \right) = \boldsymbol{C}h\left( t \right),  h'\left( t \right) = \boldsymbol{A}h\left( t \right) + \boldsymbol{B}x\left( t \right),  \\
\label{Eq1}
\end{equation}
\noindent
where the three parameters $\boldsymbol{A} \in {\mathbb{R}^{N \times N}}$, $\boldsymbol{B} \in {\mathbb{R}^N}$, and $\boldsymbol{C} \in {\mathbb{R}^N}$ represent the state matrix, input matrix, and output matrix, respectively. Since the above SSMs run on continuous inputs and are not applicable to discrete inputs such as images and text, they cannot be introduced into deep models. Thus, it is necessary to discretize them, and the zero-order hold is commonly used as a discretization method (see Appendix A). The discretized formulas are as follows
\begin{equation}
{y_t} = \boldsymbol{\bar C}{h_t}, \;\;\; {h_t} = \boldsymbol{\bar A}{h_{t - 1}} + \boldsymbol{\bar B}{x_t},   \\
\label{Eq2}
\end{equation}
\noindent
where $\boldsymbol{\bar A}$ and $\boldsymbol{\bar B}$ are the results of discretizing the continuous parameters $\boldsymbol{A}$ and $\boldsymbol{B}$ by a time scale $\boldsymbol{\rm{{\Delta }}}$, denoted as
\begin{equation}
\small
\begin{aligned}
\boldsymbol{\bar A} = {e^{\boldsymbol{{\rm{\Delta }}A}}}, \;\; \boldsymbol{\bar C} = \boldsymbol{C} , \;\;\; \boldsymbol{\bar B} = {\left( \boldsymbol{{\rm{\Delta }}A} \right)^{ - 1}} {\left( {{e^{\boldsymbol{{{\rm{\Delta }}A}}}} - \boldsymbol{I}} \right)}{\left( {\boldsymbol{{{\rm{\Delta }}B}}} \right)}.
\end{aligned}
\label{Eq3}
\end{equation}

\subsection{Mamba}
\label{Section3.2}

Since processing the input and latent state equally, previous approaches focusing on linear time-invariant SSMs (where $\boldsymbol{\bar A}$ and $\boldsymbol{\bar B}$ are invariant) may fail to capture critical information from context. Therefore, Mamba proposes a novel SSM (termed S6) by integrating an input-dependent selective mechanism into SSMs, where $\boldsymbol{\bar A}$ and $\boldsymbol{\bar B}$ are the functions of inputs, indicating Mamba is linear time-variant.

However, Mamba's linear time-variant parameters result in dynamic weights, preventing it from being computed as efficiently as linear time-invariant SSMs using the convolution. Thus, Mamba uses a parallel scanning algorithm with linear complexity \cite{17,41} to maintain efficient computation. In addition, it designs a hardware-aware algorithm by utilizing the fast SRAM in GPUs to improve efficiency.

\begin{figure*}[t]
\centering
\includegraphics[width=6.44in]{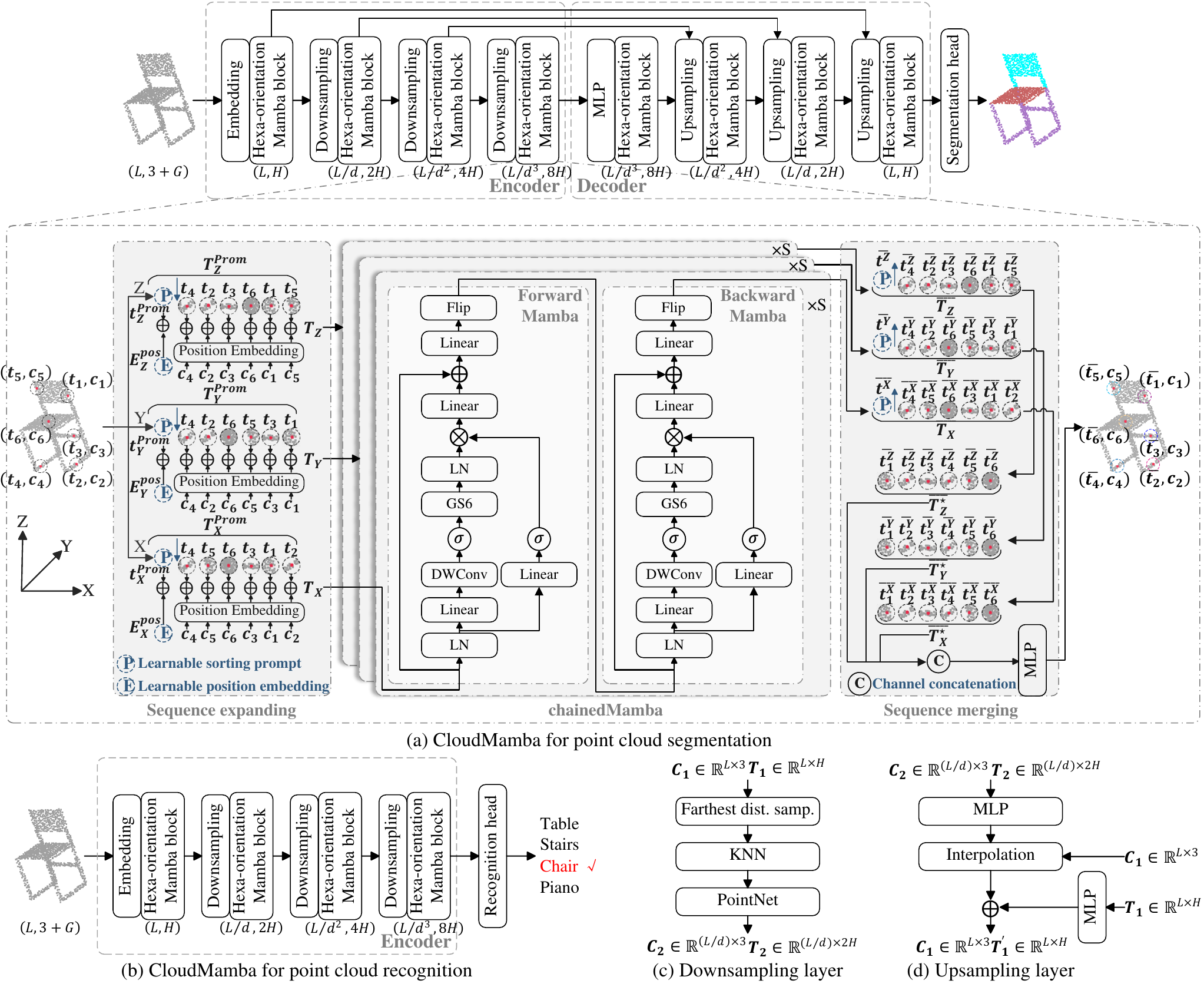}
\caption{Pipeline of our proposed network.  The Flip layer in the chainedMamba indicates a flip operation on the sequence for global modelling. S6 in Mamba is replaced by GS6. Since each point sequence is processed with forward and backward directions, there are hexa orientations for causal modeling.}
\label{Fig2}
\end{figure*}

\section{CloudMamba}
\label{Section4}

\subsection{Overview}
\label{Section4.1}

Figure \ref{Fig2} presents the overview of our network, which takes as input a point set $\boldsymbol{P} = \left\{ {\boldsymbol{{p_i}} = \left( {\boldsymbol{{c_i}},\;\boldsymbol{{f_i}}} \right)} \right\}_{i = 1}^L$, where $\boldsymbol{{c_i}} \in {\mathbb{R}^3}$ and $\boldsymbol{{f_i}} \in {\mathbb{R}^G}$ denote the 3D coordinates of $\boldsymbol{{p_i}}$ and other relevant features, respectively. In the initial stage, $\boldsymbol{P}$ is transformed into a high-dimensional space with $H$ dimensions through an embedding layer consisting of a multi-layer perceptron machine (MLP). Then, an encoder-decoder architecture built on up- and down-sampling layers and hexa-orientation Mamba blocks is used for hierarchical feature aggregation. Finally, a corresponding task head is applied. In this paper, our network is validated by recognition and segmentation results on point clouds. The recognition head processes the output of the encoder through an average pooling layer and an MLP, and the segmentation head employs an MLP to process the output of the decoder. Next, we describe components in the encoder-decoder architecture in detail.

\subsection{Hexa-orientation Mamba block}
\label{Section4.2}

The hexa-orientation Mamba block is the main feature aggregation module of our network, capable of facilitating geometry perception from multiple perspectives, and is used for features at each level to capture fine-grained structures, specifically consisting of sequence expanding, chainedMamba, and sequence merging.

\subsubsection{Sequence expanding\\}
\label{Section4.2.1}

Mamba tailored for causal sequences requires causal dependencies between elements, but visual data has a non-causal nature, such that applying Mamba directly to point clouds does not achieve expected results (see Tab. \ref{Tab5}). Existing strategies mostly build structural dependencies based on the spatial proximity of space-filling curves, yet their reliability is highly sensitive to the setting of the grid size. Hence, we build structural dependencies directly from the coordinate axis defining spatial locations. Since structural dependencies on only one axis cannot reflect rich geometric information, we construct causal sequences based on the ascending order of point coordinates $\boldsymbol{C} = \left\{ {\boldsymbol{{c_1}},\;\boldsymbol{{c_2}},\; \cdots ,\;\boldsymbol{{c_j}}} \right\}$ corresponding to input point features $\boldsymbol{T} = \left\{ {\boldsymbol{{t_1}},\;\boldsymbol{{t_2}},\; \cdots ,\;\boldsymbol{{t_j}}} \right\}$ along the Z, Y, and X axes, respectively
\begin{equation}
\boldsymbol{T_A^{Cau}} = \left\{ {\boldsymbol{{t_{{A_1}}}},\;\boldsymbol{{t_{{A_2}}}},\; \cdots ,\;\boldsymbol{{t_{{A_j}}}}} \right\}{\rm{\;}}A \in \left\{ {{\rm{Z}},{\rm{\;Y}},{\rm{\;X}}} \right\},
\label{Eq4}
\end{equation}
\noindent
where $\boldsymbol{T_A^{Cau}}$ denotes a causal sequence constructed on the $A$-axis, $\boldsymbol{t_{{A_j}}}$ represents the $j$-th point feature sorted on the $A$-axis, and ${A_j}$ corresponds to a subscript in $\boldsymbol{T}$.

\noindent
\textbf{Prompt}. Since the network involves the structural dependencies arising from three different directions (these causal sequences are essentially shared, differing only in sorting), a learnable sorting prompt $\boldsymbol{t_A^{Prom}}$ is added for each sequence to avoid confusing the network about these dependencies
\begin{equation}
{\boldsymbol{T_A^{Prom}}} = \left\{ {\boldsymbol{{t_A^{Prom}}}},\; {\boldsymbol{{t_{{A_1}}}}},\;{\boldsymbol{{t_{{A_2}}}}},\; \cdots ,\; {\boldsymbol{{t_{{A_j}}}}} \right\},
\label{Eq5}
\end{equation}
\noindent
where $\boldsymbol{t_A^{Prom}}$ is a sorting prompt added for the $A$ axis.

\noindent
\textbf{Position-aware}. Visual data is location-sensitive \cite{4}. While point coordinates contain position information, fine-grained position information may be lost in higher-level features as the network deepens. Additionally, unlike the convolution, SSMs cannot implicitly endow the network with the spatial inductive bias. Thus, we explicitly provide position information to corresponding point features by position embeddings on point coordinates as follows
\begin{equation}
\scriptsize
\boldsymbol{T_A} = \left\{ {\boldsymbol{{t_A^{Prom}}} + {\boldsymbol{E_A^{pos}}}, {\boldsymbol{{t_{{A_1}}}}} + {\boldsymbol{ \rho \left( {{c_{{A_1}}}} \right)}}, \cdots ,{\boldsymbol{{t_{{A_j}}}}} + {\boldsymbol{\rho \left( {{c_{{A_j}}}} \right)}}} \right\},
\label{Eq6}
\end{equation}
\noindent
where $\boldsymbol{E_A^{pos}}$ is a learnable position embedding we add for each sorting prompt, $\boldsymbol{c_{{A_j}}}$ is a point coordinate corresponding to the point feature $\boldsymbol{t_{{A_j}}}$, and $\rho $ is an MLP used to map point coordinates to position embeddings.

\begin{figure}[t]
\centering
\includegraphics[width=3.3in]{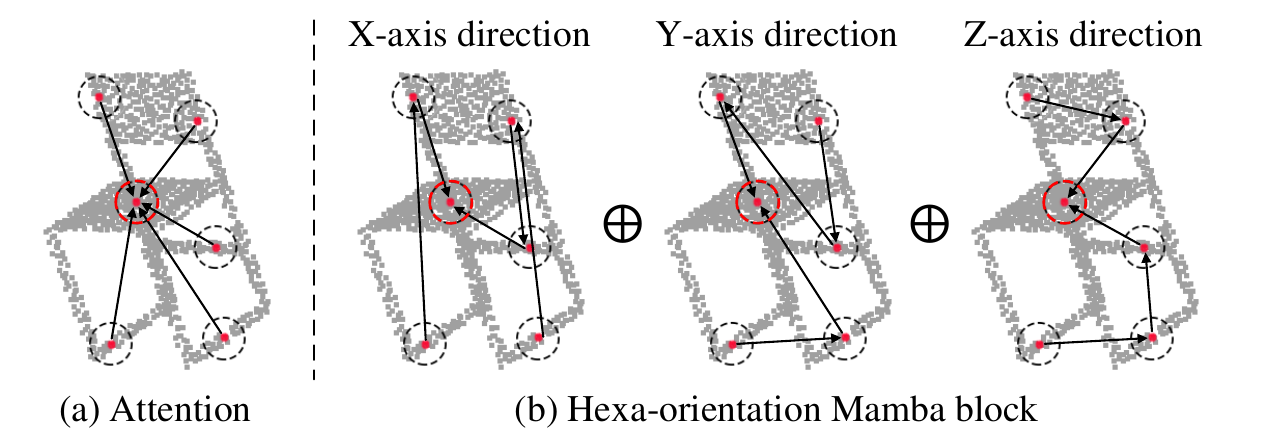}
\caption{Illustration of the global receptive fields of the attention mechanism and hexa-orientation Mamba block, with the point in the red box as an example.}
\label{Fig3}
\end{figure}

\subsubsection{ChainedMamba\\}
\label{Section4.2.2}

Although the mainstream parallel bidirectional Mamba overcomes the limitation of Mamba for visual data requiring global learning, its inference on primitive low-order geometric features leads to insufficient high-level geometric perception. Therefore, we propose chainedMamba by chaining the forward and backward processes in the parallel bidirectional Mamba, ingeniously realizing superior high-level geometric perception, as shown in Fig. \ref{Fig2}. Besides, by combining the causal sequences formed by the sequence expanding, Fig. \ref{Fig3} demonstrates that the hexa-orientation Mamba block is able to facilitate geometric learning from multiple perspectives with a global receptive field, in which the global receptive field is obtained through compressing the historical hidden state in both forward and backward directions, unlike the attention mechanism whereas the global receptive field is obtained through interactions with all elements. Although the way to obtain both global receptive fields is different, there is a close correlation between Mamba and self-attention, \textit{i.e.}, S6 is a special causal self-attention, see Appendix B.

\noindent
\textbf{GS6}. As shown in Eq. \ref{Eq1}, SSMs are sequence mappings from ${\mathbb{R}^N} \to {\mathbb{R}^N}$ on a 1-D sequence, thus S6 uses a separate set of parameters for each dimension when the input sequence has multiple dimensions, which leads to overfitting in causal inference due to excessive parameters in each dimension (see Tab. \ref{Tab4}). To improve the generalizability of the network, we propose a grouped selective state space model (GS6) through parameter sharing on S6, as shown in Algorithm 1, where ${\rm{Linea}}{{\rm{r}}_{N}}$ denotes a linear layer projected to dimension ${N}$ and ${\rm{Repeat}} \left( {\boldsymbol{V},{\rm{\;}}g} \right)$ denotes repeating an element in tensor $\boldsymbol{V}$ $g$ times. GS6 uses a grouping ratio $g$ to make every $g$ dimensions in $\boldsymbol{x}$ using the same set of parameters, alleviating the overfitting caused by S6's computational mode, while the repeating function ensures the least modification on S6's coding.
\begin{table}[t]
  \centering
  \small
\setlength{\tabcolsep}{0.1mm}{
  \begin{tabular}{l}
    \hline
    \textbf{Algorithm 1} S6 + Grouping (GS6)  \\
    \hline
    \textbf{Input}: $\boldsymbol{x}$: $\left( {B,\;L,\;D} \right)$ and a grouping rate $g$  \\
    \textbf{Output}: $\boldsymbol{y}$: $\left( {B,\;L,\;D} \right)$  \\
    1. $\boldsymbol{B}$: $\left( {B,\;L,\;N} \right)$ $ \leftarrow $ ${\rm{Linea}}{{\rm{r}}_{N}}\left( \boldsymbol{x} \right)$  \\
    2. $\boldsymbol{C}$: $\left( {B,\;L,\;N} \right)$ $ \leftarrow $ ${\rm{Linea}}{{\rm{r}}_{N}}\left( \boldsymbol{x} \right)$  \\
    /* shape of ${\bf{Paramete}}{{\bf{r}}_{\bf{\Delta }}}$ is ($D/g$), and the softplus  \\   
    ensures positive $\boldsymbol{{\rm{\Delta }}}$ */  \\
    3. $\boldsymbol{\rm{\Delta }}$: ($B, L, D/g$) $ \leftarrow $ $\log \left( {1 + {e^{\left( {{\rm{Linea}}{{\rm{r}}_{{D}/g}}\left( {\boldsymbol{x}} \right) + {\bf{Paramete}}{{\bf{r}}_{\boldsymbol{{\rm{\Delta }}}}}} \right)}}} \right)$  \\
    /* shape of ${\bf{Paramete}}{{\bf{r}}_{\boldsymbol{A}}}$ is ($D/g, N$). Each dimension \\
    in $\boldsymbol{A}$ represents a structured diagonal ${N} \times {N}$ matrix */ \\
    4. ${\boldsymbol{\bar A}}$: ($B, L, D/g, N$) $ \leftarrow $ ${\boldsymbol{\Delta}}  {\boldsymbol{\otimes}}  {\bf{Paramete}}{{\bf{r}}_{\boldsymbol{A}}}$  \\
    5. ${\boldsymbol{\bar B}}$: ($B, L, D/g, N$) $ \leftarrow $ ${\boldsymbol{\Delta}}  {\boldsymbol{\otimes}}  \boldsymbol{B}$  \\
    6. ${\boldsymbol{\bar A}},\;{\boldsymbol{\bar B}}$: ($B, L, D, N$) $ \leftarrow $ Repeat(${\boldsymbol{\bar A}},\;g$), Repeat(${\boldsymbol{\bar B}},{\rm{\;}}g$)  \\
    /* time-variant: calculated by the parallel scanning */  \\
    7. $\boldsymbol{y}$: ($B, L, D$) $ \leftarrow $ SSM$\left( {{\boldsymbol{\bar A}},\;{\boldsymbol{\bar B}},\;{\boldsymbol{C}}} \right)\left( {\boldsymbol{x}} \right)$  \\
    return $\boldsymbol{y}$  \\
    \hline
  \end{tabular}}
\end{table}

\subsubsection{Sequence merging\\}
\label{Section4.2.3}

After the chainedMamba process different causal sequences ${\left\{ {\boldsymbol{{T_A}},\;A \in \left\{ {{\rm{Z}},{\rm{\;Y}},{\rm{\;X}}} \right\}} \right\}} $, these sequences $ \left( \left\{ {{\boldsymbol{\overline {{T_A}}}}  = \left\{ {{\boldsymbol{\overline {{t^A}}}} ,\;{\boldsymbol{\overline {t_{{A_1}}^A}}} ,\; \cdots ,\;{\boldsymbol{\overline {t_{{A_j}}^A} }}} \right\}} \right\} \right) $ contain the higher-order interaction features causally inferred from different perspectives, respectively, and the sequence merging is responsible for integrating them together to captures rich geometric information. Specifically, the sorting prompts (the first term) in the sequences are first removed and the sequences are returned to their original order $\left( {\left\{ {\boldsymbol{\overline {T_A^ \star }}  = \left\{ {\boldsymbol{\overline {t_1^A}} ,\;{\boldsymbol{\overline {t_2^A}}} ,\; \cdots ,\;{\boldsymbol{\overline {t_j^A} }}} \right\}} \right\}} \right)$, and then the sequences are merged by Eq. \ref{Eq7}, where ${\rm{Concat}}$ denotes channel-wise concatenation on the corresponding element and $\gamma $ is an MLP scaled down by three times, keeping the output dimension of the hexa-orientation Mamba block the same as the input dimension.
\begin{equation}
\boldsymbol{Output} = \gamma \left( {{\rm{Concat}}\left( {\left\{ {\boldsymbol{\overline {T_A^ \star }} ,\;A \in \left\{ {{\rm{Z}},{\rm{\;Y}},{\rm{\;X}}} \right\}} \right\}} \right)} \right)
\label{Eq7}
\end{equation}

\subsection{Downsampling and upsampling layers}
\label{Section4.3}

\noindent
\textbf{Downsampling}. The downsampling layer is to reduce the cardinality of point sets at a downsampling rate in the encoder. Assume that the input point coordinate set and its corresponding point feature set of the downsampling layer are $\boldsymbol{{C_1}} \in {\mathbb{R}^{L \times 3}}$ and $\boldsymbol{{T_1}} \in {\mathbb{R}^{L \times H}}$, respectively, and the downsampling rate is $d$. First, the farthest distance sampling \cite{42} is performed in $\boldsymbol{C_1}$ to determine a point coordinate subset $\boldsymbol{{C_2}} \in {\mathbb{R}^{\left( {L/d} \right) \times 3}}$ with $L/d$ sampled points. Then, to pool the point features on $\boldsymbol{{C_1}}$ to $\boldsymbol{{C_2}}$, we perform the $K$-nearest neighbor on $\boldsymbol{{C_1}}$ for each point coordinate in $\boldsymbol{{C_2}}$ and pass the neighboring point features of each point in $\boldsymbol{{C_2}}$ through PointNet \cite{43} to construct a point feature set $\boldsymbol{{T_2}} \in {\mathbb{R}^{\left( {L/d} \right) \times 2H}}$ corresponding to $\boldsymbol{{C_2}}$. The overview of the downsampling layer is shown in Fig. \ref{Fig2}(c).

\noindent
\textbf{Upsampling}. For dense tasks such as point cloud segmentation, we follow the U-Net design \cite{44}, where the above encoder is coupled with a symmetric decoder. The upsampling layer serves as a connection between successive stages in the decoder, and it is to recover the cardinality of point sets according to the corresponding encoder stage. Suppose that the input point coordinate set and its corresponding point feature set of the upsampling layer are $\boldsymbol{{C_2}} \in {\mathbb{R}^{\left( {L/d} \right) \times 3}}$ and $\boldsymbol{{T_2}} \in {\mathbb{R}^{\left( {L/d} \right) \times 2H}}$, respectively, and the point coordinate set and its point feature set of the corresponding encoder stage are $\boldsymbol{{C_1}} \supset \boldsymbol{{C_2}}$ ($\boldsymbol{{C_1}} \in {\mathbb{R}^{L \times 3}}$) and $\boldsymbol{{T_1}} \in {\mathbb{R}^{L \times H}}$, respectively. First, the dimension of $\boldsymbol{{T_2}}$ is aligned to $\boldsymbol{{T_1}}$ by an MLP (\textit{i.e.}, $\boldsymbol{{T_2}} \in {\mathbb{R}^{\left( {L/d} \right) \times H}}$), and the tri-linear interpolation method is performed on $\boldsymbol{{C_2}}$ for each point in $\boldsymbol{{C_1}}$ to obtain the interpolated point features corresponding to $\boldsymbol{{C_1}}$ from $\boldsymbol{{T_2}}$. Then, $\boldsymbol{{T_1}}$ passing through an MLP is summed with these interpolated point features by the skip connection to obtain a point feature set $\boldsymbol{T_1^\prime}  \in {\mathbb{R}^{L \times H}}$ corresponding to $\boldsymbol{{C_1}}$ at the decoder stage. Fig. \ref{Fig2}(d) illustrates the structure of the upsampling layer.

\section{Experiment}
\label{Section5}

To validate CloudMamba's potential in point cloud analysis, it is compared with different types of networks on ModelNet40, ScanObjectNN, ShapeNet, and S3DIS datasets. For a detailed introduction on datasets and evaluation metrics, see Appendix C. Moreover, we explore the network property and the impact of the components on the performance by extensive ablation comparisons. Also, the FLOPs (floating point operations) and Params reported, reflecting computation and space complexity, respectively, are measured on the same RTX 4090 GPU to ensure a fair comparison.

\subsection{Point cloud recognition}
\label{Section5.2}

Table \ref{Tab1} compares the experimental results of our network and different operators-based mainstream works on ModelNet40 dataset. As shown in the fifth column in Tab. \ref{Tab1}, CloudMamba achieves a higher OA compared to the concurrent SSM-based works, which indicates that our network is able to take full advantage of Mamba's superior long-range modeling. Besides, CloudMamba achieves competitive accuracy with the state-of-the-art Point Transformer but consuming less computational resources, which not only stresses the strong geometric representation capability of our network, but also proves it improves computational efficiency based on the linear complexity of Mamba. To further validate the robustness of the network to complex scenarios, the sixth column in Tab. \ref{Tab1} lists the experimental results of our network and different operators-based works on challenging ScanObjectNN (PB\_T50\_RS) dataset, where CloudMamba outperforms the state-of-the-art PointConT and PCM, which strongly proves its robustness. 

\subsection{Point cloud part segmentation}
\label{Section5.3}

Table \ref{Tab2} compares the experimental results of our network with different operators-based mainstream works on ShapeNet dataset. As shown in Tab. \ref{Tab2}, our network achieves a higher Instance mIoU, and significantly improves the accuracy compared to the concurrent SSM-based works. It is worth noting that CloudMamba achieves the same Instance mIoU as the state-of-the-art Point Transformer with fewer FLOPs and Params, which fully validates our network is capable of adapting Mamba's superior long-range modeling to point cloud analysis. Also, we observe CloudMamba requires only 1.234G FLOPs, significantly lower than the attention networks and  other SSM networks, which demonstrates our network enables a lightweight architecture.

\begin{table}[t]
  \centering
  \small
\setlength{\tabcolsep}{0.4mm}{
  \begin{tabular}{lccccc}
    \hline
\multirow{2}{*}{Networks} & \multirow{2}{*}{Operators} & \multirow{2}{*}{FLOPs} & \multirow{2}{*}{Params} & \multicolumn{2}{c}{OA}  \\
& & & & Model. &  Scan. \\
    \hline
PointNet++ \shortcite{42} & MLP & 4.1 & 1.7 & 91.9 & 77.9 \\
PointMLP \shortcite{45} & MLP & 15.7 & 13.2 & 93.6 & 85.4 \\
PointNext \shortcite{46} & MLP & 1.6 & 1.4 & 93.2 & 87.7 \\
    \hline
DGCNN \shortcite{50} & CNN & - & - & 92.9 & 78.1 \\
3D-GCN \shortcite{48} & CNN & - & - & 92.1 & - \\
    \hline
Point Transformer \shortcite{4} & Attention & 18.4 & 9.6 & \textbf{93.7} & - \\
Point-BERT \shortcite{56} & Attention & 2.3 & 22.1 & 92.7 & 83.1  \\
Point-MAE \shortcite{57} & Attention & 2.4 & 22.1 & 93.2 & 85.2  \\
OctFormer \shortcite{10} & Attention & 0.6 & 3.4 & 92.7 & - \\
IDPT \shortcite{68} & Attention & 2.3 & 1.7 & 93.4 & 83.7 \\
PointGST \shortcite{70} & Attention & 4.8 & 0.6 & 93.4 & 85.6  \\
DAPT \shortcite{69} & Attention & 5.0 & 1.1 & 93.1 & 85.4 \\
LCM \shortcite{71} & Attention & 1.3 & 2.7 & 93.6 & 87.8 \\
    \hline
PointTramba \shortcite{58} & \makecell{SSM \& \\ Attention} & - & - & 92.7 & - \\
Mamba3D \shortcite{37} & SSM & 3.9 & 16.9 & 93.4 & 87.6 \\
PointMamba \shortcite{38} & SSM & 3.6 & 12.3 & 92.4 & 82.5  \\
OctMamba \shortcite{39} & SSM & 1.3 & 3.1 & 92.7 & -\\
PCM \shortcite{40} & SSM & 45.0 & 34.2 & 92.6 & 88.1 \\
SI-Mamba \shortcite{73} & SSM & 3.6 & 12.3 & 92.7 & 87.3 \\ 
CloudMamba & SSM & 1.150 & 9.95 & \textbf{93.7} & \textbf{88.3} \\
    \hline
  \end{tabular}}
  \caption{Experimental results of our network and different operators-based mainstream works on ModelNet40 and ScanObjectNN datasets. }
  \label{Tab1}
\end{table}

\begin{table}[t]
  \centering
  \small
\setlength{\tabcolsep}{0.8mm}{
  \begin{tabular}{lcccc}
    \hline
Networks & Operator & FLOPs  & Params & Ins. mIoU  \\
    \hline
PointNet++ \shortcite{42} & MLP & 4.8 & 1.7 & 85.1  \\
PointMLP \shortcite{45} & MLP & 6.3 & 16.8 & 86.1  \\
ReCon \shortcite{61} & MLP & - & - & 86.4  \\
    \hline
DGCNN \shortcite{50} & CNN & - & - & 85.2  \\
3D-GCN \shortcite{48} & CNN & - & - & 85.3  \\
    \hline
Point Transformer \shortcite{4} & Attention & 36.7 & 19.4 & \textbf{86.6}  \\
PCT \shortcite{25} & Attention & 4.4 & 2.9 & 86.4  \\
Point-BERT \shortcite{56} & Attention & 4.5 & 22.1 & 85.6  \\
Point-MAE \shortcite{57} & Attention & 4.8 & 22.1 & 86.1  \\
MaskPoint \shortcite{60} & Attention & 4.8 & 22.1 & 86.0  \\
APES \shortcite{47} & Attention & - & - & 85.8  \\
IDPT \shortcite{68} & Attention & 4.8 & 5.7 & 85.7 \\
LCM \shortcite{71} & Attention & - & - & 86.3 \\
PointGST \shortcite{70} & Attention & 4.8 & 5.6 & 85.7  \\
DAPT \shortcite{69} & Attention & 5.0 & 5.7 & 85.5 \\
    \hline
PointTramba \shortcite{58} & \makecell{SSM \& \\ Attention} & - & - & 85.7  \\
Mamba3D \shortcite{37} & SSM & 11.8 & 23.0 & 85.6  \\
PointMamba \shortcite{38} & SSM & 14.3 & 17.4 & 85.8  \\
PCM \shortcite{40} & SSM & 52.1 & 40.6 & 84.3  \\
PMA \shortcite{75} & SSM & - & - & 86.1  \\
CloudMamba & SSM & 1.234 & 16.57 & \textbf{86.6}  \\
    \hline
  \end{tabular}}
  \caption{Experimental results of our network and different operators-based mainstream works on ShapeNet dataset.}
  \label{Tab2}
\end{table}

\begin{table}[!t]
  \centering
  \small
\setlength{\tabcolsep}{0.75mm}{
  \begin{tabular}{ccccc}
    \hline
Networks & Operator & FLOPs  & Params & mIoU  \\
    \hline
RandLA-Net \shortcite{66} & MLP & - & - & 63.2  \\
PointMeta \shortcite{65} & MLP & - & - & 69.5  \\
\hline
DGCNN \shortcite{50} & CNN & - & - & 56.5  \\
3D-GCN \shortcite{48} & CNN &	- & - & 58.6  \\
\hline
Point Transformer \shortcite{4} & Attention &	147.2 & 19.4 & 70.4  \\
Point Transformer v2 \shortcite{5} & Attention &	88.3 & 12.9 & 71.6  \\
SuperpointTransformer \shortcite{67} & Attention & - & - & 68.9  \\
Point Transformer v3 \shortcite{12} & Attention &	26.5 & 46.2 & 73.4  \\
\hline
Grid Mamba \shortcite{76} & SSM & - & - & 71.8 \\
Pamba \shortcite{72} & SSM & - & - & 73.5  \\
PCM \shortcite{40} & SSM & 72.1 & 40.6 & 63.4  \\
CloudMamba & SSM & 4.922 & 16.56 & \textbf{73.6} \\
\hline
  \end{tabular}}
  \caption{Experimental results of our network and different operators-based mainstream works on S3DIS dataset.}
  \label{Tab3}
\end{table}

\subsection{Point cloud semantic segmentation}
\label{Section5.4}

Table \ref{Tab3} compares the experimental results of our network and different operators-based mainstream works on S3DIS dataset. As shown in Tab. \ref{Tab3}, fewer SSM-based networks are validated on difficult scene-level S3DIS dataset, while CloudMamba makes up for this deficiency, and further demonstrates our network's exceptional long-range modeling capability and efficient linear complexity property by outperforming the state-of-the-art Point Transformer V3 with fewer FLOPs and params.

\subsection{Ablation studies}
\label{Section5.5}

\subsubsection{Validity checking\\}
\label{Section5.5.1}

In the proposed network, we design the sorting prompt, position embedding, and GS6 to further enhance the network performance. To demonstrate their validity, we compare the experimental results with and without these components, where the sorting prompt and position embedding are directly removed from the network when they are not used, and S6 is used as a replacement when GS6 is not used (\textit{i.e.}, the grouping rate is set to 1 in GS6).

\noindent
\textbf{Sorting prompt}. As shown in Group I in Tab. \ref{Tab4}, the sorting prompt improves 0.46\% OA, indicating that as the network continuously learns, the causal dependencies generated from different directions are confused, leading to incorrect shape understanding, and the sorting prompt, as a widget, can help the network clearly infer geometric structures from different dependencies with little effect on the FLOPs and Params.

\noindent
\textbf{Position embedding}. Group II in Tab. \ref{Tab4} shows that the absence of the position embedding decreases 0.5\% OA, which indicates that as the network deepens, higher-level features lose fine position information. Hence, it is necessary to facilitate geometric learning by continuously maintaining position information through the position embedding.

\noindent
\textbf{GS6}. Group III in Tab. \ref{Tab4} shows GS6 improves 1.54\% OA while reducing 0.4M params, revealing the computational mode in S6 causes a certain degree of overfitting, and GS6 is able to alleviate this problem via the parameter sharing.

\begin{table}[t]
  \centering
  \small
\setlength{\tabcolsep}{5mm}{
  \begin{tabular}{c|ccc}
    \hline
 & FLOPs (G) & Params (M) & OA (\%)  \\
    \hline
I & \multicolumn{3}{c}{Sorting prompt}  \\
w/ & 1.150 & 9.95 & \textbf{93.65}  \\
w/o & 1.150 & 9.95 &  93.19  \\
    \hline
II & \multicolumn{3}{c}{Position embedding}  \\
w/ & 1.150 & 9.95 & \textbf{93.65}  \\
w/o & 1.141 & 9.70 & 93.15  \\
    \hline
III & \multicolumn{3}{c}{GS6}  \\
w/ & 1.150 & 9.95 & \textbf{93.65}  \\
w/o & 1.150 & 10.35 &  92.11 \\
    \hline
  \end{tabular}}
  \caption{Experimental results of our network with and without the relevant components on ModelNet40 dataset.}
  \label{Tab4}
\end{table}

\subsubsection{Network property\\}
\label{Section5.5.2}

\begin{table}[!t]
  \centering
  \small
\setlength{\tabcolsep}{3.5mm}{
  \begin{tabular}{cccc}
    \hline
 Combination & FLOPs (G) & Params (M) & OA (\%)  \\
    \hline
None & 1.123 & 4.99 & 88.98  \\
X & 1.123 & 4.99 & 91.37  \\
Y & 1.123 & 4.99 & 90.96  \\
Z & 1.123 & 4.99 & 91.09  \\
X, Y & 1.136 & 7.47 & 92.38  \\	
X, Z & 1.136 & 7.47 & 92.71 \\	
Y, Z & 1.136 & 7.47 & 92.59 \\	
X, Y, Z & 1.150 & 9.95 & \textbf{93.65}  \\
    \hline
  \end{tabular}}
  \caption{Experimental results in combinations of the sequences from Z, Y, and X axes on ModelNet40 dataset, where None denotes point sets are not causally sorted.}
  \label{Tab5}
\end{table}

\begin{table}[!t]
  \centering
  \small
\setlength{\tabcolsep}{4mm}{
  \begin{tabular}{cccc}
    \hline
Structure & FLOPs (G) & Params (M) & OA (\%)  \\
    \hline
Parallel & 1.150 & 9.95 & 92.69 \\
Chained & 1.150 & 9.95 & \textbf{93.65}  \\
    \hline
  \end{tabular}}
  \caption{Experimental results with different structures of the bidirectional Mamba on ModelNet40 dataset.}
  \label{Tab6}
\end{table}

\noindent
\textbf{Causal dependency}. Our network forms  three causal sequences, providing multiple perspectives for geometric perception. To explore the impact of these sequences, we perform separate experiments in combinations of the sequences from Z, Y, and X axes, as shown in Tab. \ref{Tab5}, where  more causal sequences involved are able to obtain a higher OA, which reveals more causal dependencies from multiple perspectives can help the network to learn stereoscopic geometry. Notably, no causal sorting of point sets (input directly) achieves a low OA, which reflects the causal nature of Mamba.

\noindent
\textbf{Structure of the bidirectional Mamba}. Existing Mamba-based visual networks mostly follow the idea of Zhu \textit{et al.} \cite{31} to adapt Mamba's unidirectional modeling to visual data, \textit{i.e.}, the parallel bidirectional Mamba, but it results in insufficient high-level geometric perception compared to our chainedMamba. We compare the experimental results of this parallel structure with the chained structure adopted by our network. For a fair comparison, the experimental configurations are identical (including the number of the bidirectional Mamba, and the use of GS6) except for employing different structures of the bidirectional Mamba. Tab. \ref{Tab6} shows our chained structure obtains a higher OA, which means our structure is able to achieve more superior geometric perception and yield better expressivity.

\section{Conclusion}
\label{Section6}

In this paper, we propose an SSM-based point cloud network termed CloudMamba, which overcomes the shortcomings of existing related works, including imperfect point cloud serialization, insufficient high-level geometric perception, and overfitting caused by the computational mode in S6, through the sequence expanding \& merging, chainedMamba, and GS6. Extensive experiments demonstrate CloudMamba is able to obtain state-of-the-art results with linear complexity.

\section*{Acknowledgments}

This work is supported in part by the National Natural Science Foundation of China under Grant (62476126, 62272227).

\bibliography{aaai2026}

@inproceedings{1,
author = "A. Vaswani and N. Shazeer and N. Parmar and J. Uszkoreit and L. Jones and A. N. Gomez and L. Kaiser and I. Polosukhin",
title = "{Attention is all you need}",
booktitle = "NIPS",
address = "Long Beach, CA, USA",
pages = "5998--6008",
month = "Dec.",
year = 2017
}

@inproceedings{2,
author = "A. Dosovitskiy and L. Beyer and A. Kolesnikov and D. Weissenborn and X. H. Zhai and T. Unterthiner and M. Dehghani and M. Minderer and G. Heigold and S. Gelly and J. Uszkoreit and N. Houlsby",
title = "{An image is worth 16x16 words: Transformers for image recognition at scale}",
booktitle = "ICLR",
address = "Virtual Event, Austria",
pages = "1--22",
month = "May",
year = 2021
}

@inproceedings{3,
author = "X. Yan and C. D. Zheng and Z. Li and S. Wang and S. G. Cui",
title = "{Pointasnl: Robust point clouds processing using nonlocal neural networks with adaptive sampling}",
booktitle = "CVPR",
address = "Virtual Event / Seattle, WA, USA",
pages = "5588--5597",
month = "Jun.",
year = 2020
}

@inproceedings{4,
author = "H. S. Zhao and L. Jiang and J. Y. Jia and P. Torr and V. Koltun",
title = "{Point transformer}",
booktitle = "ICCV",
address = "Montreal, BC, Canada",
pages = "16239--16248",
month = "Oct.",
year = 2021
}

@inproceedings{5,
author = "X. Y. Wu and Y. X. Lao and L. Jiang and X. H. Liu and H. S. Zhao",
title = "{Point transformer V2: Grouped vector attention and partition-based pooling}",
booktitle = "NIPS",
address = "New Orleans, LA, USA",
pages = "1--13",
month = "Dec.",
year = 2022
}

@inproceedings{6,
author = "D. Nie and R. Lan and L. Wang and X. F. Ren",
title = "{Pyramid Architecture for Multi-Scale Processing in Point Cloud Segmentation}",
booktitle = "CVPR",
address = "New Orleans, LA, USA",
pages = "17263--17273",
month = "Jun.",
year = 2022
}

@inproceedings{7,
author = "L. Fan and Z. Q. Pang and T. Y. Zhang and Y. X. Wang and H. Zhao and F. Wang and N. Y. Wang and Z. X. Zhang",
title = "{Embracing single stride 3D object detector with sparse transformer}",
booktitle = "CVPR",
address = "New Orleans, LA, USA",
pages = "8448--8458",
month = "Jun.",
year = 2022
}

@inproceedings{9,
author = "C. Park and Y. Jeong and M. S. Cho and J. Park",
title = "{Fast point transformer}",
booktitle = "CVPR",
address = "New Orleans, LA, USA",
pages = "16928--16937",
month = "Jun.",
year = 2022
}

@article{10,
author = {P. S. Wang},
title = {OctFormer: Octree-based Transformers for 3D Point Clouds},
journal = {TOG},
volume = {42},
number = {4},
pages = {155},
month = {Jul.},
year = {2023}
}

@inproceedings{11,
author = "Z. J. Liu and X. Y. Yang and H. T. Tang and S. Yang and S. Han",
title = "{FlatFormer: Flattened Window Attention for Efficient Point Cloud Transformer}",
booktitle = "CVPR",
address = "Vancouver, BC, Canada",
pages = "1200--1211",
month = "Jun.",
year = 2023
}

@inproceedings{12,
author = "X. Y. Wu and L. Jiang and P. S. Wang and Z. J. Liu and X. H. Liu and Y. Qiao and W. L. Ouyang and T. He and H. S. Zhao",
title = "{Point Transformer V3: Simpler, Faster, Stronger}",
booktitle = "CVPR",
address = "Seattle, WA, USA",
pages = "1--15",
month = "Jun.",
year = 2024
}

@inproceedings{13,
author = "A. Gu and I. Johnson and K. Goel and K. Saab and T. Dao and A. Rudra and C. Ré",
title = "{Combining recurrent, convolutional, and continuous-time models with linear state space layers}",
booktitle = "NIPS",
address = "Virtual Event",
pages = "572--585",
month = "Dec.",
year = 2021
}

@inproceedings{14,
author = "A. Gu and T. Dao and S. Ermon and A. Rudra and C. Ré",
title = "{Hippo: Recurrent memory with optimal polynomial projections}",
booktitle = "NIPS",
address = "Virtual Event",
pages = "1474--1487",
month = "Dec.",
year = 2020
}

@inproceedings{15,
author = "A. Gu and K. Goel and C. Ré",
title = "{Efficiently modeling long sequences with structured state spaces}",
booktitle = "ICLR",
address = "Virtual Event",
pages = "1--15",
month = "Apr.",
year = 2022
}

@misc{16,
 author = {H. Mehta and A. Gupta and A. Cutkosky and B. Neyshabur},
 title = {Long range language modeling via gated state spaces},
 eprint = {2206.13947},
 archivePrefix = {arXiv},
 month = {Jun.},
 year = {2022}
}

@misc{17,
 author = {J. T. H. Smith and A. Warrington and S. W. Linderman},
 title = {Simplified state space layers for sequence modeling},
 eprint = {2208.04933},
 archivePrefix = {arXiv},
 month = {Aug.},
 year = {2022}
}

@inproceedings{18,
author = "A. Gupta and A. Gu and J. Berant",
title = "{Diagonal State Spaces are as Effective as Structured State Spaces}",
booktitle = "NIPS",
address = "New Orleans, LA, USA",
pages = "1--12",
month = "Dec.",
year = 2022
}

@misc{19,
 author = {A. Gu and T. Dao},
 title = {Mamba: Linear-time sequence modeling with selective state spaces},
 eprint= {2312.00752},
 archivePrefix={arXiv},
 month = {Dec.},
 year = {2023}
}

@inproceedings{20,
author = "T. Dao and D. Y. Fu and S. Ermon and A. Rudra and C. Ré",
title = "{FlashAttention: Fast and Memory-Efficient Exact Attention with IO-Awareness}",
booktitle = "NIPS",
address = "New Orleans, LA, USA",
pages = "1--16",
month = "Dec.",
year = 2022
}

@inproceedings{22,
author = "Z. R. Wu and S. R. Song and A. Khosla and F. Yu and L. G. Zhang and X. O. Tang and J. X. Xiao",
title = "{3d shapenets: A deep representation for volumetric shapes}",
booktitle = "CVPR",
address = "Boston, MA, USA",
pages = "1912--1920",
month = "Jun.",
year = 2015
}

@inproceedings{23,
author = "M. A. Uy and Q. H. Pham and B. S. Hua and D. T. Nguyen and S. K. Yeung",
title = "{Revisiting Point Cloud Classification: A New Benchmark Dataset and Classification Model on Real-World Data}",
booktitle = "ICCV",
address = "Seoul, South Korea",
pages = "1588--1597",
month = "Oct.",
year = 2019
}

@article{24,
author = {L. Yi and V. G. Kim and D. Ceylan and I. C. Shen and M. Y. Yan and H. Su and C. Lu and Q. X. Huang and A. Sheffer and L. Guibas},
title = {A scalable active framework for region annotation in 3d shape collections},
journal = {TOG},
volume = {35},
number = {6},
pages = {210},
month = {Dec.},
year = {2016}
}

@article{25,
author = {M. H. Guo and J. X. Cai and Z. N. Liu and T. J. Mu and R. R. Martin and S. M. Hu},
title = {PCT: Point cloud transformer},
journal = {Computational Visual Media},
volume = {7},
number = {2},
pages = {187--199},
month = {Apr.},
year = {2021}
}

@inproceedings{29,
author = "D. Hilbert",
title = "{Über die stetige abbildung einer linie auf ein flächenstück}",
booktitle = "Dritter Band: Analysis· Grundlagen der Mathematik· Physik Verschiedenes: Nebst Einer Lebensgeschichte",
year = 1935
}

@misc{30,
 author = {Y. Liu and Y. J. Tian and Y. Z. Zhao and H. T. Yu and L. X. Xie and Y. W. Wang and Q. X. Ye and Y. F. Liu},
 title = {VMamba: Visual State Space Model},
 eprint = {2401.10166},
 archivePrefix= {arXiv},
 month = {Jan.},
 year = {2024}
}

@inproceedings{31,
author = "L. H. Zhu and B. C. Liao and Q. Zhang and X. L. Wang and W. Y. Liu and X. G. Wang",
title = "{Vision Mamba: Efficient Visual Representation Learning with Bidirectional State Space Model}",
booktitle = "ICML",
address = "Vienna, Austria",
pages = "1--10",
month = "Jul.",
year = 2024
}

@misc{33,
 author = {J. C. Ruan and S. C. Xiang},
 title = {VM-UNet: Vision Mamba UNet for Medical Image Segmentation},
 eprint = {2402.02491},
 archivePrefix= {arXiv},
 month = {Feb.},
 year = {2024}
}

@misc{35,
 author = {J. Ma and F. F. Li and B. Wang},
 title = {U-Mamba: Enhancing Long-range Dependency for Biomedical Image Segmentation},
 eprint = {2401.04722},
 archivePrefix= {arXiv},
 month = {Jan.},
 year = {2024}
}

@misc{36,
 author = {C. Wang and O. Tsepa and J. Ma and B. Wang},
 title = {Graph-Mamba: Towards Long-Range Graph Sequence Modeling with Selective State Spaces},
 eprint = {2402.00789},
 archivePrefix= {arXiv},
 month = {Feb.},
 year = {2024}
}

@inproceedings{37,
author = "X. Han and Y. Tang and Z. X. Wang and X. Z. Li",
title = "{Mamba3D: Enhancing Local Features for 3D Point Cloud Analysis via State Space Model}",
booktitle = "ACM International Conference on Multimedia",
address = "Melbourne, Australia",
pages = "1--10",
month = "Oct.",
year = 2024
}

@misc{38,
author = {D. K. Liang and X. Zhou and W. Xu and X. K. Zhu and Z. K. Zou and X. Q. Ye and X. Tan and X. Bai},
title = {PointMamba: A Simple State Space Model for Point Cloud Analysis},
 eprint = {2402.10739},
 archivePrefix= {arXiv},
 month = {Feb.},
 year = {2024}
}

@misc{39,
 author = {J. M. Liu and R. J. Yu and Y. Wang and Y. Zheng and T. C. Deng and W. C. Ye and H. S. Wang},
 title = {Point Mamba: A Novel Point Cloud Backbone Based on State Space Model with Octree-Based Ordering Strategy},
 eprint = {2403.06467},
 archivePrefix= {arXiv},
 month = {Mar.},
 year = {2024}
}

@misc{40,
 author = {T. Zhang and X. T. Li and H. B. Yuan and S. P. Ji and S. C. Yan},
 title = {Point Cloud Mamba: Point Cloud Learning via State Space Model},
 eprint = {2403.00762},
 archivePrefix= {arXiv},
 month = {May},
 year = {2024}
}

@inproceedings{41,
author = "E. Martin and C. Cundy",
title = "{Parallelizing linear recurrent neural nets over sequence length}",
booktitle = "ICLR",
address = "Vancouver, BC, Canada",
pages = "1--9",
month = "Apr.",
year = 2018
}

@inproceedings{42,
author = "C. R. Qi and L. Yi and H. Su and L. J. Guibas",
title = "{Pointnet++: Deep hierarchical feature learning on point sets in a metric space}",
booktitle = "NIPS",
address = "Long Beach, CA, USA",
pages = "5099--5108",
month = "Dec.",
year = 2017
}

@inproceedings{43,
author = "C. R. Qi and H. Su and K. C. Mo and L. J. Guibas",
title = "{Pointnet: Deep learning on point sets for 3d classification and segmentation}",
booktitle = "CVPR",
address = "Honolulu, HI, USA",
pages = "77--85",
month = "Jul.",
year = 2017
}

@inproceedings{44,
author = "O. Ronneberger and P. Fischer and T. Brox",
title = "{U-Net: Convolutional Networks for Biomedical Image Segmentation}",
booktitle = "International Conference on Medical Image Computing and Computer-Assisted Intervention",
address = "Munich, Germany",
pages = "234--241",
month = "Oct.",
year = 2015
}

@inproceedings{45,
author = "X. Ma and C. Qin and H. X. You and H. X. Ran and Y. Fu",
title = "{Rethinking Network Design and Local Geometry in Point Cloud: A Simple Residual MLP Framework}",
booktitle = "ICLR",
address = "Virtual Event",
pages = "1--14",
month = "Apr.",
year = 2022
}

@inproceedings{46,
author = "G. C. Qian and Y. C. Li and H. W. Peng and J. J. Mai and H. A. A. K. Hammoud and M. Elhoseiny and B. Ghanem",
title = "{PointNeXt: Revisiting PointNet++ with Improved Training and Scaling Strategies}",
booktitle = "NIPS",
address = "New Orleans, LA, USA",
pages = "1--13",
month = "Dec.",
year = 2022
}

@inproceedings{47,
author = "C. Z. Wu and J. W. Zheng and J. Pfrommer and J. Beyerer",
title = "{Attention-Based Point Cloud Edge Sampling}",
booktitle = "CVPR",
address = "Vancouver, BC, Canada",
pages = "5333--5343",
month = "Jun.",
year = 2023
}

@article{48,
author = {Z. H. Lin and S. Y. Huang and Y. F. Wang},
title = {Learning of 3D Graph Convolution Networks for Point Cloud Analysis},
journal = {PAMI},
volume = {44},
number = {8},
pages = {4212--4224},
month = {Aug.},
year = 2022
}

@article{50,
author = {Y. Wang and Y. B. Sun and Z. W. Liu and S. E. Sarma and M. M. Bronstein and J. M. Solomon},
title = {Dynamic Graph CNN for Learning on Point Clouds},
journal = {TOG},
volume = {38},
number = {5},
pages = {146},
month = {Oct.},
year = {2019}
}

@inproceedings{56,
author = "X. M. Yu and L. L. Tang and Y. M. Rao and T. J. Huang and J. Zhou and J. W. Lu",
title = "{Point-BERT: Pre-training 3D Point Cloud Transformers with Masked Point Modeling}",
booktitle = "CVPR",
address = "New Orleans, LA, USA",
pages = "19291--19300",
month = "Jun.",
year = 2022
}

@inproceedings{57,
author = "Y. T. Pang and W. X. Wang and F. E. H. Tay and W. Liu and Y. H. Tian and L. Yuan",
title = "{Masked Autoencoders for Point Cloud Self-supervised Learning}",
booktitle = "ECCV",
address = "Tel Aviv, ISRAEL",
pages = "604--621",
month = "Oct.",
year = 2022
}

@misc{58,
 author = {Z. C. Wang and Z. H. Chen and Y. M. Wu and Z. Zhao and L. P. Zhou and D. Xu},
 title = {PoinTramba: A Hybrid Transformer-Mamba Framework for Point Cloud Analysis},
 eprint = {2405.15463},
 archivePrefix= {arXiv},
 month = {May},
 year = {2024}
}

@inproceedings{60,
author = "H. T. Liu and M. Cai and Y. J. Lee",
title = "{Masked discrimination for self-supervised learning on point clouds}",
booktitle = "ECCV",
address = "Tel Aviv, Israel",
pages = "657--675",
month = "Oct.",
year = 2022
}

@inproceedings{61,
author = "Z. K. Qi and R. P. Dong and G. F. Fan and Z. Ge and X. Y. Zhang and K. S. Ma and L. Yi",
title = "{Contrast with Reconstruct: Contrastive 3D Representation Learning Guided by Generative Pretraining}",
booktitle = "ICML",
address = "Honolulu, HI, USA",
pages = "28223--28243",
month = "Jul.",
year = 2023
}

@inproceedings{63,
author = "I. Armeni and O. Sener and A. R. Zamir and H. L. Jiang and I. Brilakis and M. Fischer and S. Savarese",
title = "{3D semantic parsing of large-scale indoor spaces}",
booktitle = "CVPR",
address = "Las Vegas, NV, USA",
pages = "1534--1543",
month = "Jun.",
year = 2016
}

@inproceedings{65,
author = "H. J. Lin and X. W. Zheng and L. J. Li and F. Chao and S. S. Wang and Y. Wang and Y. H. Tian and R. R. Ji",
title = "{Meta Architecture for Point Cloud Analysis}",
booktitle = "CVPR",
address = "Vancouver, BC, Canada",
pages = "17682--17691",
month = "Jun.",
year = 2023
}

@article{66,
author = {Q. Y. Hu and B. Yang and L. H. Xie and S. Rosa and Y. L. Guo and Z. H. Wang and N. Trigoni and A. Markham},
title = {Learning Semantic Segmentation of Large-Scale Point Clouds With Random Sampling},
journal = {PAMI},
volume = {44},
number = {11},
pages = {8338--8354},
month = {May},
year = {2022}
}

@inproceedings{67,
author = "D. Robert and H. Raguet and L. Landrieu",
title = "{Efficient 3d semantic segmentation with superpoint transformer}",
booktitle = "ICCV",
address = "Paris, France",
pages = "1--10",
month = "Oct.",
year = 2023
}

@inproceedings{68,
author = "Y. H. Zha and J. P. Wang and T. Dai and B. Chen and Z. Wang and S. T. Xia",
title = "{Instance-aware Dynamic Prompt Tuning for Pre-trained Point Cloud Models}",
booktitle = "ICCV",
address = "Paris, France",
pages = "1--10",
month = "Oct.",
year = 2023
}

@inproceedings{69,
author = "X. Zhou and D. K. Liang and W. Xu and X. K. Zhu and Y. H. Xu and Z. K. Zou and X. Bai",
title = "{Dynamic Adapter Meets Prompt Tuning: Parameter-Efficient Transfer Learning for Point Cloud Analysis}",
booktitle = "CVPR",
address = "Seattle, WA, USA",
pages = "14707--14717",
month = "Jun.",
year = 2024
}

@misc{70,
 author = {D. K. Liang and T. R. Feng and X. Zhou and Y. M. Zhang and Z. K. Zou and X. Bai},
 title = {Parameter-Efficient Fine-Tuning in Spectral Domain for Point Cloud Learning},
 eprint = {2410.08114},
 archivePrefix= {arXiv},
 month = {Oct.},
 year = {2024}
}

@inproceedings{71,
author = "Y. H. Zha and N. Q. Li and Y. Z. Wang and T. Dai and H. Guo and B. Chen and Z. Wang and Z. H. Ouyang and S. T. Xia",
title = "{LCM: Locally Constrained Compact Point Cloud Model for Masked Point Modeling}",
booktitle = "NIPS",
address = "Vancouver, BC, Canada",
pages = "1--14",
month = "Dec.",
year = 2024
}

@Inproceedings{72,
author = "Z. Y. Li and Y. B. Ai and J. H. Lu and C. X. Wang and J. C. and Deng and H. Z. Chang and Y. Z. Liang and W. F. Yang and S. F. Zhang and T. Z. Zhang",
title = "{Pamba: Enhancing Global Interaction in Point Clouds via State Space Model}",
booktitle = "AAAI",
address = "Philadelphia, PA, USA",
pages = "5092--5100",
month = "Feb.",
year = 2025
}

@Inproceedings{73,
author = "A. Bahri and M. Yazdanpanah and M. Noori and S. Dastani and M. Cheraghalikhani and D. Osowiechi and G. Hakim and F. Beizaee and I. B. Ayed and C. Desrosiers",
title = "{Spectral Informed Mamba for Robust Point Cloud Processing}",
booktitle = "CVPR",
address = "Nashville, TN, USA",
pages = "11799--11809",
month = "Jun.",
year = 2025
}

@Inproceedings{74,
author = "J. M. Liu and J. R. Han and L. H. Liu and A. I. Aviles-Rivero and C. K. Jiang and Z. Liu and H. S. Wang",
title = "{Mamba4D: Efficient 4D Point Cloud Video Understanding with Disentangled Spatial-Temporal State Space Models}",
booktitle = "CVPR",
address = "Nashville, TN, USA",
pages = "17626--17636",
month = "Jun.",
year = 2025
}

@Inproceedings{75,
author = "Y. H. Zha and Y. Z. Wang and H. Guo and J. P. Wang and T. Dai and B. Chen and Z. H. Ouyang and Y. R. Xue and K. Chen and S. T. Xia",
title = "{PMA: Towards Parameter-Efficient Point Cloud Understanding via Point Mamba Adapter}",
booktitle = "CVPR",
address = "Nashville, TN, USA",
pages = "16976--16986",
month = "Jun.",
year = 2025
}

@article{76,
author = {Y. L. Yang and T. Z. Xun and K. R. Hao and B. Wei and X. S. Tang},
title = {Grid Mamba:Grid State Space Model for Large-Scale Point Cloud Analysis},
journal = {Neurocomputing},
volume = {636},
pages = {129985},
month = {Jul.},
year = {2025}
}

\clearpage
\fullwidthtitle{CloudMamba: Grouped Selective State Spaces for Point Cloud Analysis \\ Supplementary Material}

\section*{Appendix A Discretization of continuous SSMs}
\label{App1}

In this section, we deduce in detail the process of discretizing continuous SSMs by the zero-order holding method.

The state equation of continuous SSMs is as follows
\begin{equation}
h'\left( t \right) = {\boldsymbol{A}}h\left( t \right) + {\boldsymbol{B}}x\left( t \right).
\label{Eq8}
\end{equation}

\noindent
Simultaneously multiplied by ${e^{ - {\boldsymbol{A}}t}}$:
\begin{equation}
\begin{aligned}
{e^{ - {\boldsymbol{A}}t}}h'\left( t \right) = {e^{ - {\boldsymbol{A}}t}}{\boldsymbol{A}}h\left( t \right) + \;{e^{ - {\boldsymbol{A}}t}}{\boldsymbol{B}}x\left( t \right),  \\
{e^{ - {\boldsymbol{A}}t}}h'\left( t \right) - {e^{ - {\boldsymbol{A}}t}}{\boldsymbol{A}}h\left( t \right) = \;{e^{ - {\boldsymbol{A}}t}}{\boldsymbol{B}}x\left( t \right),  \\
\frac{{d\left[ {{e^{ - {\boldsymbol{A}}t}}h\left( t \right)} \right]}}{{dt}} = \;{e^{ - {\boldsymbol{A}}t}}{\boldsymbol{B}}x\left( t \right). \;\;\;\;\;\;\;\;
\end{aligned}
\label{Eq9}
\end{equation}

\noindent
Followed by simultaneous integrals:
\begin{equation}
\small
\begin{aligned}
\mathop \smallint \nolimits_{{t_{\left( {k - 1} \right)}}}^{{t_k}} d\left[ {{e^{ - {\boldsymbol{A}}t}}h\left( t \right)} \right] = \mathop \smallint \nolimits_{{t_{\left( {k - 1} \right)}}}^{{t_k}} {e^{ - {\boldsymbol{A}}t}}{\boldsymbol{B}}x\left( t \right)dt, \;\;\;\;\;\;\;\;\;\  \\
{e^{ - {\boldsymbol{A}}{t_k}}}h\left( {{t_k}} \right) - {e^{ - {\boldsymbol{A}}{t_{\left( {k - 1} \right)}}}}h\left( {{t_{\left( {k - 1} \right)}}} \right) = \mathop \smallint \nolimits_{{t_{\left( {k - 1} \right)}}}^{{t_k}} {e^{ - {\boldsymbol{A}}\tau }}{\boldsymbol{B}}x\left( \tau  \right)d\tau ,  \\
{e^{ - {\boldsymbol{A}}{t_k}}}h\left( {{t_k}} \right) = {e^{ - {\boldsymbol{A}}{t_{\left( {k - 1} \right)}}}}h\left( {{t_{\left( {k - 1} \right)}}} \right) + \mathop \smallint \nolimits_{{t_{\left( {k - 1} \right)}}}^{{t_k}} {e^{ - {\boldsymbol{A}}\tau }}{\boldsymbol{B}}x\left( \tau  \right)d\tau,  \\
h\left( {{t_k}} \right) = {e^{{\boldsymbol{A}}\left( {{t_k} - {t_{\left( {k - 1} \right)}}} \right)}}h\left( {{t_{\left( {k - 1} \right)}}} \right) + \mathop \smallint \nolimits_{{t_{\left( {k - 1} \right)}}}^{{t_k}} {e^{{\boldsymbol{A}}\left( {{t_k} - \tau } \right)}}{\boldsymbol{B}}x\left( \tau  \right)d\tau,  \\
h\left( {{t_k}} \right) = {e^{\Delta {\boldsymbol{A}}}}h\left( {{t_{\left( {k - 1} \right)}}} \right) + \mathop \smallint \nolimits_{{t_{\left( {k - 1} \right)}}}^{{t_k}} {e^{{\boldsymbol{A}}\left( {{t_k} - \tau } \right)}}{\boldsymbol{B}}x\left( \tau  \right)d\tau. \;\;\;\;\;\;\;  \\
\end{aligned}
\label{Eq10}
\end{equation}

\noindent
Let $\nu  = {t_k} - \tau $, then $d\nu  =  - d\tau $. Since the zero-order holding makes $x\left( t \right)$ a constant ${x_k}$, Eq. \ref{Eq10} is simplified to:
\begin{equation}
\small
\begin{aligned}
h\left( {{t_k}} \right) = {e^{\Delta {\boldsymbol{A}}}}h\left( {{t_{\left( {k - 1} \right)}}} \right) + \mathop \smallint \nolimits_{{t_{\left( {k - 1} \right)}}}^{{t_k}} {e^{{\boldsymbol{A}}\left( {{t_k} - \tau } \right)}}{\boldsymbol{B}}x\left( \tau  \right)d\tau   \\
 = {e^{\Delta {\boldsymbol{A}}}}h\left( {{t_{\left( {k - 1} \right)}}} \right) - \mathop \smallint \nolimits_\Delta ^0 {e^{{\boldsymbol{A}}\nu }}d\nu {\boldsymbol{B}}{x_k} \;\;\;\;\;\;\;\;\;\;\;\;\;\;\;\;\;\;  \\
 = {e^{\Delta {\boldsymbol{A}}}}h\left( {{t_{\left( {k - 1} \right)}}} \right) + \mathop \smallint \nolimits_0^\Delta  {e^{{\boldsymbol{A}}\nu }}d\nu {\boldsymbol{B}}{x_k} \;\;\;\;\;\;\;\;\;\;\;\;\;\;\;\;\;  \\
 = {e^{\Delta {\boldsymbol{A}}}}h\left( {{t_{\left( {k - 1} \right)}}} \right) + \frac{{\left( {{e^{\Delta {\boldsymbol{A}}}} - {\boldsymbol{I}}} \right)\Delta }}{{{\boldsymbol{A}}\Delta }}{\boldsymbol{B}}{x_k},\;\;\;\;\;\;\;\;\;\;  \\
\end{aligned}
\label{Eq11}
\end{equation}

\noindent
By writing the above in a discrete form:
\begin{equation}
\begin{aligned}
{h_k} = {e^{\Delta {\boldsymbol{A}}}}{h_{k - 1}} + {\left( {\Delta {\boldsymbol{A}}} \right)^{ - 1}}\left( {{e^{\Delta {\boldsymbol{A}}}} - {\boldsymbol{I}}} \right)\left( {\Delta {\boldsymbol{B}}} \right){x_k},  \\
{h_k} = {\boldsymbol{\bar A}}{h_{k - 1}} + {\boldsymbol{\bar B}}{x_k}, \;\;\;\;\;\;\;\;\;\;\;\;\;\;\;\;\;\;\;\;\;\;\;
\end{aligned}
\label{Eq12}
\end{equation}

\noindent
where ${\boldsymbol{\bar A}} = {e^{\Delta {\boldsymbol{A}}}}$ and ${\boldsymbol{\bar B}} = {\left( {\Delta {\boldsymbol{A}}} \right)^{ - 1}}\left( {{e^{\Delta {\boldsymbol{A}}}} - {\boldsymbol{I}}} \right)\left( {\Delta {\boldsymbol{B}}} \right)$.

\section*{Appendix B Relationship between S6 (Mamba) and the self-attention}
\label{App2}

For an input sequence ${\boldsymbol{x}} = \left\{ {{x_1},\;{x_2},\; \cdots ,\;{x_L}} \right\} \in {{\mathbb{R}}^{L \times D}}$, S6 regards $\boldsymbol{\Delta} $, $\boldsymbol{B}$, and $\boldsymbol{C}$ in Eq. \ref{Eq3} as the functions of $\boldsymbol{x}$. Therefore, $\boldsymbol{\Delta _t}$, $\boldsymbol{B_t}$, and $\boldsymbol{C_t}$ in the $t$-th time step is computed as follows
\begin{equation}
\begin{aligned}
{\boldsymbol{{\Delta _t}}} = \log \left( {1 + {e^{{\rm{Linea}}{{\rm{r}}^\Delta }\left( {{x_t}} \right)}}} \right),  \\
{\boldsymbol{{B_t}}} = {\rm{Linea}}{{\rm{r}}^B}\left( {{x_t}} \right), \;\;\;\;\;\;\;  \\
{\boldsymbol{{C_t}}} = {\rm{Linea}}{{\rm{r}}^C}\left( {{x_t}} \right), \;\;\;\;\;\;\;  \\
\end{aligned}
\label{Eq13}
\end{equation}

\noindent
where ${\rm{Linea}}{{\rm{r}}^\Delta }$ denotes a linear projection used for $\Delta$. Taking Eq. \ref{Eq13} into Eq. \ref{Eq3} yields $\boldsymbol{\Delta _t}$, $\boldsymbol{\overline {{A_t}}} $, $\boldsymbol{\overline {{B_t}}} $, and $\boldsymbol{{C_t}}$. Next, we unfold Eq. \ref{Eq2}, then the output ${\boldsymbol{y}} = \left\{ {{y_1},\;{y_2},\; \cdots ,\;{y_L}} \right\} \in {\mathbb{R}^{L \times D}}$ is computed as follows
\begin{equation}
\begin{aligned}
{h_t} = \mathop \sum \limits_{i = 1}^t \left( {\mathop \prod \limits_{j = i + 1}^t {\boldsymbol{\overline {{A_j}}}} } \right){\boldsymbol{\overline {{B_i}}}} {x_i},  \\
{y_t} = {\boldsymbol{{C_t}}}{h_t}. \;\;\;\;\;\;\;\;\;\;\;\;\;\;
\end{aligned}
\label{Eq14}
\end{equation}

\noindent
${h_t}$ in Eq. \ref{Eq14} can be expressed in a matrix form:
\begin{equation}
\left[ {\begin{array}{*{20}{c}}
{{h_1}}\\
{{h_2}}\\
 \vdots \\
{{h_t}}
\end{array}} \right] = \left[ {\begin{array}{*{20}{c}}
{\overline {\boldsymbol{B_1}} }&0& \cdots &0\\
{\overline {\boldsymbol{A_2}} \;\overline {\boldsymbol{B_1}} }&{\overline {\boldsymbol{B_2}} }& \cdots &0\\
 \vdots & \vdots & \ddots &0\\
{\mathop \prod \limits_{j = 2}^t \overline {\boldsymbol{A_j}} \overline {\boldsymbol{B_1}} }&{\mathop \prod \limits_{j = 3}^t \overline {\boldsymbol{A_j}} \overline {\boldsymbol{B_2}} }& \cdots &{\overline {\boldsymbol{B_t}} }
\end{array}} \right]\left[ {\begin{array}{*{20}{c}}
{{x_1}}\\
{{x_2}}\\
 \vdots \\
{{x_t}}
\end{array}} \right].
\label{Eq15}
\end{equation}

\noindent
As a result, the transformation matrix $\boldsymbol{W}$ from $\boldsymbol{x}$ to $\boldsymbol{y}$ can be expressed as
\begin{equation}
\begin{aligned}
{\boldsymbol{W_{i,\;j}}} = {\boldsymbol{{C_i}}}\left( {\mathop \prod \limits_{s = j + 1}^i {\boldsymbol{\overline {{A_s}}}} } \right){\boldsymbol{\overline {{B_j}}}} \;\;\;\;\;\;\;\;\;\;\;\;\;\;\;\;\;\;\;\;\;\;  \\
 = {\boldsymbol{{C_i}}}\left( {\mathop \prod \limits_{s = j + 1}^i {e^{\boldsymbol{{\Delta _s}A}}}} \right){\boldsymbol{\overline {{B_j}}}} \;\;\;\;\;\;\;\;\;\;\;\;\;\;\;\;\;\; \\
 = {\boldsymbol{{C_i}}}{e^{\left( {\mathop \sum \limits_{s = j + 1}^i {\boldsymbol{{\Delta _s}A}}} \right)}}{\boldsymbol{\overline {{B_j}}}} \;\;\;\;\;\;\;\;\;\;\;\;\;\;\;\;\;\;\;\;\;\;\; \\ 
 = {\boldsymbol{{C_i}}}{e^{\left( {\mathop \sum \limits_{s = j + 1}^i \log \left( {1 + {e^{{\rm{Linea}}{{\rm{r}}^\Delta }\left( {{x_s}} \right)}}} \right){\boldsymbol{A}}} \right)}}{\boldsymbol{\overline {{B_j}}}}.  \\
\end{aligned}
\label{Eq16}
\end{equation}

\noindent
Let ${\boldsymbol{{Q_i}}} = {\boldsymbol{{C_i}}}$, ${\boldsymbol{{K_j}}} = {\boldsymbol{{\left( {\overline {{B_j}} } \right)^T}}}$, and ${{\boldsymbol{H_{i,\;j}}}} = {e^{\left( {\mathop \sum \limits_{s = j + 1}^i \log \left( {1 + {e^{{\rm{Linea}}{{\rm{r}}^\Delta }\left( {{x_s}} \right)}}} \right){\boldsymbol{A}}} \right)}}$, then Eq. \ref{Eq16} can be abbreviated as

\begin{equation}
{\boldsymbol{{W_{i,\;j}}}} = {\boldsymbol{{Q_i}{H_{i,\;j}}K_j^T}}.
\label{Eq17}
\end{equation}

\noindent
Eq. \ref{Eq17} indicates S6 captures the effects of ${x_i}$ and ${x_j}$ by $\boldsymbol{Q_i}$ and $\boldsymbol{K_j^T}$, respectively, and that $\boldsymbol{H_{i,\;j}}$ contains causal information from ${x_i}$ to ${x_j}$. Note that $\boldsymbol{W_{i,\;j}}$ is similar to the attention mechanism with a mask $\boldsymbol{M}$ (a lower triangular matrix with elements set to 1), \textit{i.e.}, the causal self-attention, which reveals that S6 is a special causal self-attention modulated by $\boldsymbol{H_{i,\;j}}$. Moreover, it is known from Eq. \ref{Eq16} that the causal information contained in $\boldsymbol{H_{i,\;j}}$ is directly affected by $\boldsymbol{A}$, which explains the importance of designing the Hippo algorithm by Gu \textit{et al.} \cite{14} to obtain more accurate causal information.

\section*{Appendix C Datasets and evaluation metrics}
\label{App3}

ModelNet40 is a well-segmented synthetic recognition dataset containing 12,311 samples with 40 categories, of which 9,843 samples are used for training and 2,468 samples used for testing. Following the data preprocessing of Qi \textit{et al.} \cite{43}, 1024 points (normalized to a unit sphere) and their corresponding normal vectors are uniformly sampled. As per most relevant studies in the literature, we adopt the overall accuracy (OA) as an evaluation metric.

ScanObjectNN is a challenging recognition dataset from the real world. This dataset introduces several variants through the perturbation. We conduct experiments on the most difficult variant, PB\_T50\_RS, to validate the robustness of CloudMamba, where the suffixes T50, R, and S denote randomly shifting the bounding box by 50\% along each axis, rotation, and scaling, respectively. This variant contains about 15,000 valid samples with 15 categories, where 80\% of the samples are used for training and 20\% of the samples are used for testing. We use the same data preprocessing and evaluation metric as on ModelNet40 dataset.

ShapeNet is a large-scale part segmentation dataset containing 16,878 samples with 50 parts from 16 categories, of which 14,005 samples are used for training and 2,874 samples are used for testing. We adopt the same data preprocessing as on ModelNet40 dataset and use the Instance mIoU (Ins. mIoU) as an evaluation metric according to  general practice in the community.

S3DIS is a scene-level semantic segmentation dataset that includes 3D scanned points from 271 rooms in 6 regions, with 13 semantic label categories. We follow the data preprocessing procedure of Qi \textit{et al.} \cite{43} and take point coordinates, RGB, and normalized position as inputs. As per most relevant studies in the literature, region 5 is used as test set and the other regions are used for training, and the mean IoU (mIoU) is used as an evaluation metric.

\section*{Appendix D Effect of the grouping rate}
\label{App4}

Group III in Tab. \ref{Tab4} demonstrates that GS6 can alleviate a certain degree of overfitting caused by S6 through the parameter sharing, where the grouping rate parameter determines how many dimensions share the same set of parameters in GS6. Therefore, we continuously increase the grouping rate to explore its effect on the network. Tab. \ref{Tab13} reveals that the overfitting of S6 is gradually alleviated as the grouping rate increases, but it leads to underfitting when it is overlarge. The grouping rate is a hyperparameter similar to the number of heads in the multi-head self-attention. Hence, it is important to set an appropriate grouping rate for the performance of GS6 based on the model scale, and when deploying the model to another data, we recommend setting the grouping rate starting from 3 and gradually increasing it until performance degradation is observed.

\begin{table}[h]
  \centering
  \small
\setlength{\tabcolsep}{3.4mm}{
  \begin{tabular}{cccc}
    \hline
Grouping rate & FLOPs (G) & Params (M) & OA (\%)  \\
    \hline
1 & 1.150 & 10.35 & 93.11  \\
2 & 1.150 & 10.05 & 93.35 \\
3†† & 1.150 & 9.95 & \textbf{93.65}  \\
6 & 1.150 & 9.85 & 92.75 \\
9 & 1.150	&	9.82 & 92.14\\
    \hline
  \end{tabular}}
  \caption{Experimental results of our network with different grouping rates on ModelNet40 dataset. †† indicates a baseline parameter.}
  \label{Tab13}
\end{table}

\section*{Appendix E Comparison with other SSM-based networks}
\label{App5}

\begin{figure}[t]
\centering
\includegraphics[width=2.8in]{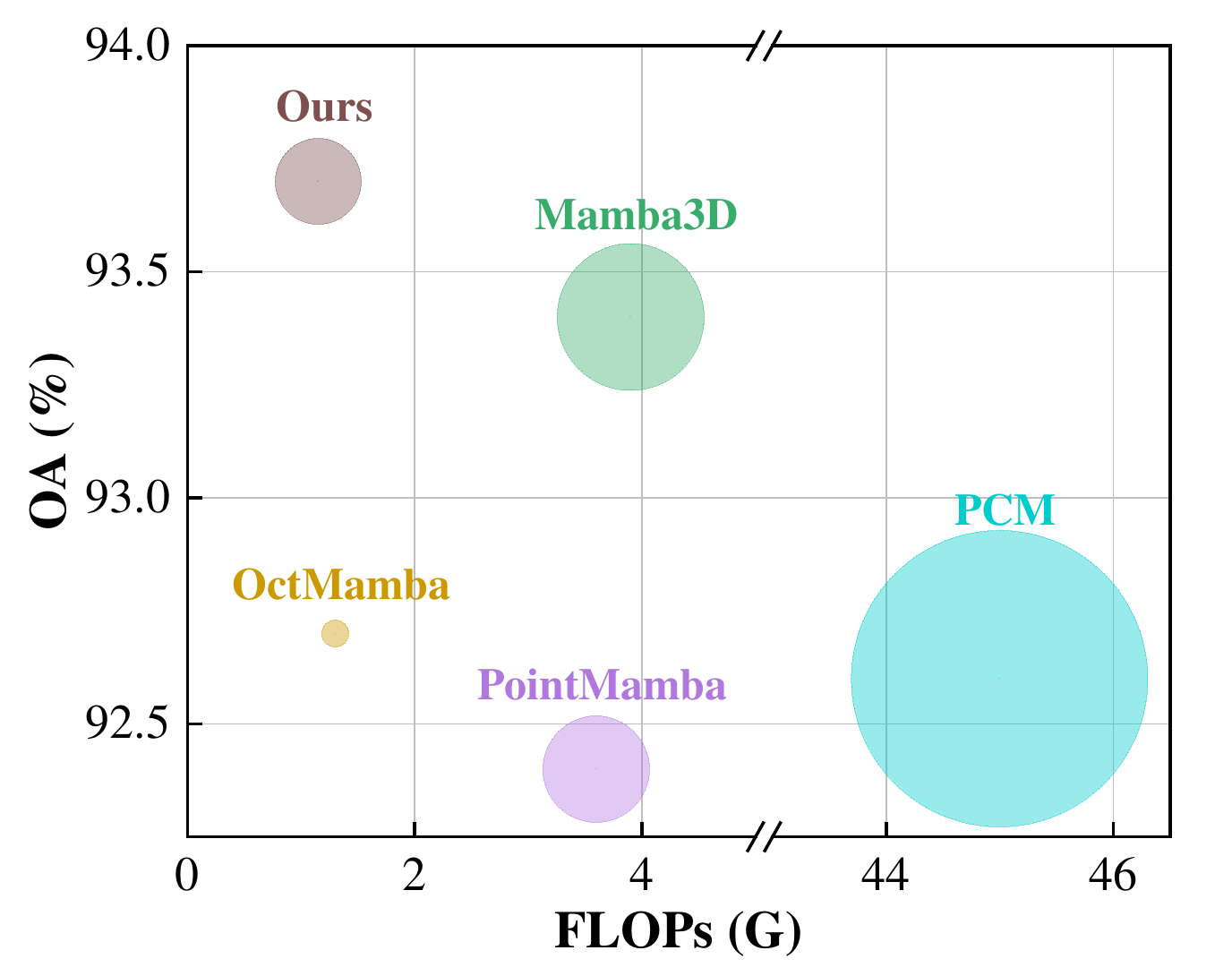}
\caption{Comparison of SSM-based networks on ModelNet40 dataset, where a larger circle means more parameters.}
\label{Fig5}
\end{figure}

To intuitively demonstrate that our CloudMamba is able to achieve state-of-the-art results with less computational resources in contemporaneous SSM-based networks, we visualize the accuracy of each SSM network on ModelNet40 dataset, as well as the FLOPs and Params. As shown in Fig. \ref{Fig5}, our network achieve state-of-the-art accuracy with less FLOPs and Params among SSM networks.

\section*{Appendix F Point cloud serialization}
\label{App6}

It is crucial to construct structural dependencies in a sequence via point cloud serialization for Mamba's causal inference. At present, space-filling curves are one of main serialization strategies, among which the Hilbert curve and its variant Trans-Hilbert are commonly used methods in the community. Hence, we apply both to the causal dependency construction of our network to replace the sequence expanding module. Since the grid size is a core parameter of space-filling curves for restricting each point to the corresponding grid, we set different grid sizes for the Hilbert curve. The comparison in Tab. \ref{Tab0} demonstrates that space-filling curves are highly sensitive to the grid size, while our proposed sequence expanding is simple yet effective.


\section*{Appendix G Linear complexity}
\label{App7}

\begin{figure}[t]
\centering
\includegraphics[width=2.8in]{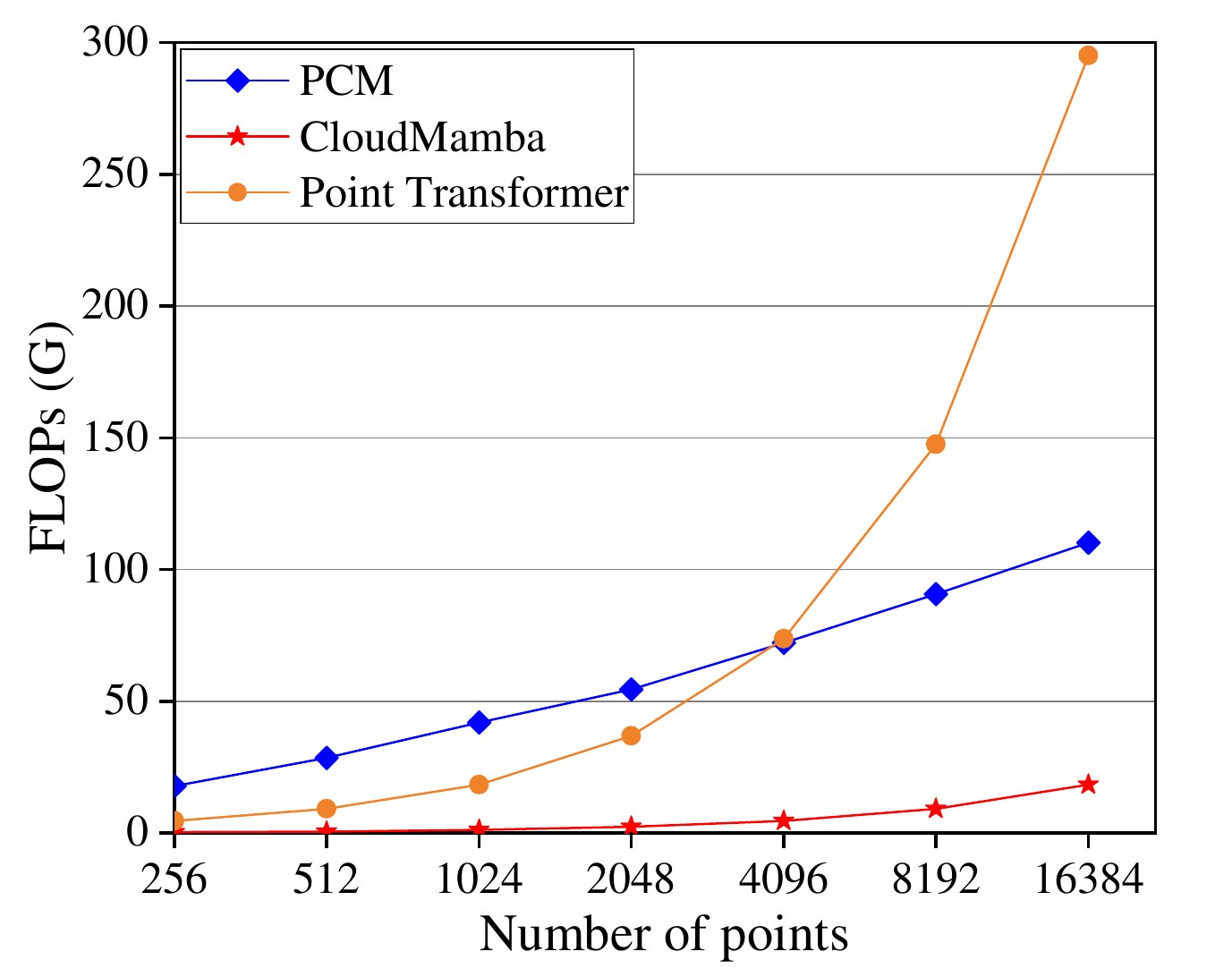}
\caption{Changes of different networks on the FLOPs as the number of points increases.}
\label{Fig4}
\end{figure}

\noindent
Mamba achieves efficient computation by the parallel scanning algorithm with linear complexity. To validate our network inherits this property, we continuously increase the number of input points to observe changes on the FLOPs, while adding Point Transformer \cite{4} and PCM \cite{40} as a contrast, as shown in Fig. \ref{Fig4}, which demonstrates the feasibility of our network for long sequence modeling in terms of computational overheads.

\section*{Appendix H Limitations}
\label{App8}

\begin{table*}[t]
  \centering
  \small
\setlength{\tabcolsep}{0.72mm}{
  \begin{tabular}{lcccccccccccccccc}
    \hline
Networks & \makecell{Airplane \\ \#341} & \makecell{Bag	 \\ \#14} & \makecell{Cap \\ \#11} & \makecell{Car \\ \#158} & \makecell{Chair \\ \#704} & \makecell{Earphone \\ \#14} & \makecell{Guitar \\ \#159} & \makecell{Knife \\ \#80} & \makecell{Lamp \\ \#286} & \makecell{Laptop \\ \#83} & \makecell{Motorbike \\ \#51} & \makecell{Mug \\ \#38} & \makecell{Pistol \\ \#44} & \makecell{Rocket \\ \#12} & \makecell{Skateboard \\ \#31} & \makecell{Table \\ \#848}  \\
    \hline
PointNet++ & 82.4 & 79.0 & 87.7 & 77.3 & 90.8 & 71.8 & 91.0 & 85.9 & 83.7 & 95.3 & 71.6 & 94.1 & 81.3 & 58.7 & 76.4 & 82.6  \\
PointNet & 83.4 & 78.7 & 82.5 & 74.9 & 89.6 & 73.0 & 91.5 & 85.9 & 80.8 & 95.3 & 65.2 & 93.0 & 81.2 & 57.9 & 72.8 & 80.6  \\
PointMLP & 83.5 & 83.4 & 87.5 & 80.5 & 90.3 & 78.2 & \textbf{92.2} & \textbf{88.1} & 82.6 & 96.2 & \textbf{77.5} & 95.8 & \textbf{85.4} & 64.6 & \textbf{83.3} & 84.3  \\
3D-GCN & 82.8 & \textbf{86.1} & 84.8 & 79.2 & 91.1 & 74.9 & 91.6 & 87.4 & 83.6 & 95.8 & 69.3 & 94.9 & 82.4 & 61.1 & 75.6 & 82.2  \\
DGCNN & 84.0 & 83.4 & 86.7 & 77.8 & 90.6 & 74.7 & 91.2 & 87.5 & 82.8 & 95.7 & 66.3 & 94.9 & 81.1 & 63.5 & 74.5 & 82.6  \\
SpiderCNN & 83.5 & 81.0 & 87.2 & 77.5 & 90.7 & 76.8 & 91.1 & 87.3 & 83.3 & 95.8 & 70.2 & 93.5 & 82.7 & 59.7 & 75.8 & 82.8  \\
PointASNL & 84.1 & 84.7 & 87.9 & 79.7 & \textbf{92.2} & 73.7 & 91.0 & 87.2 & 84.2 & 95.8 & 74.4 & 95.2 & 81.0 & 63.0 & 76.3 & 83.2  \\
PCT & 85.0 & 82.4 & 89.0 & \textbf{81.2} & 91.9 & 71.5 & 91.3 & \textbf{88.1} & \textbf{86.3} & 95.8 & 64.6 & 95.8 & 83.6 & 62.2 & 77.6 & 83.7  \\
ACT & \textbf{85.2} & 85.2 & 88.8 & \textbf{81.2} & 91.3 & 79.4 & \textbf{92.2} & 87.9 & 85.8 & 96.0 & 75.5 & 95.5 & 85.2 & \textbf{66.6} & 77.7 & 81.5  \\
Point-BERT & 84.3 & 84.8 & 88.0 & 79.8 & 91.0 & \textbf{81.7} & 91.6 & 87.9 & 85.2 & 95.6 & 75.6 & 94.7 & 84.3 & 63.4 & 76.3 & 81.5  \\
Point-MAE & 84.3 & 85.0 & 88.3 & 80.5 & 91.3 & 78.5 & 92.1 & 87.4 & 86.1 & 96.1 & 75.2 & 94.6 & 84.7 & 63.5 & 77.1 & 82.4  \\
MaskPoint & 84.2 & 85.6 & 88.1 & 80.3 & 91.2 & 79.5 & 91.9 & 87.8 & 86.2 & 95.3 & 76.9 & 95.0 & 85.3 & 64.4 & 76.9 & 81.8  \\
CloudMamba & 84.5 & 83.3 & \textbf{91.7} & 79.2 & \textbf{92.2} & 72.9 & 91.7 & 87.9 & 84.8 & \textbf{96.5} & 59.6 & \textbf{96.2} & 82.2 & 62.7 & 74.2 & \textbf{84.5}  \\
    \hline
  \end{tabular}}
  \caption{mIoU on each category in ShapeNet dataset by CloudMamba and comparison networks.}
  \label{Tab9}
\end{table*}

\begin{table*}[t]
  \centering
  \small
\setlength{\tabcolsep}{3mm}{
  \begin{tabular}{cccccccccc}
    \hline
Airplane & Bathtub & Bed & Bench & Bookshelf & Bottle & Bowl & Car & Chair & Cone  \\
\#100 & \#50 & \#100 & \#20 & \#100 & \#100 & \#20 & \#100 & \#100 & \#20  \\
    \hline
1.00 & 0.94 & 1.00 & 0.80 & 0.97 & 0.97 & 0.90 & 0.99 & 0.97 & 1.00  \\
    \hline
Cup & Curtain & Desk & Door & Dresser & Flower\_pot & Glass\_box & Guitar & Keyboard & Lamp  \\
\#20 & \#20 & \#86 & \#20 & \#86 & \#20 & \#100 & \#100 & \#20 & \#20  \\
    \hline
0.70 & 0.90 & 0.87 & 0.90 & 0.81 & 0.55 & 0.96 & 0.99 & 1.00 & 0.95  \\
    \hline
Laptop & Mantel & Monitor & Night\_stand & Person & Piano & Plant & Radio & Range\_hood & Sink  \\
\#20 & \#100 & \#100 & \#86 & \#20 & \#100 & \#100 & \#20 & \#100 & \#20  \\
    \hline
1.00 & 0.96 & 0.99 & 0.92 & 1.00 & 0.91 & 0.89 & 0.90 & 0.92 & 0.90  \\
    \hline
Sofa & Stairs & Stool & Table & Tent & Toilet & Tv\_stand & Vase & Wardrobe & Xbox  \\
\#100 & \#20 & \#20 & \#100 & \#20 & \#100 & \#100 & \#100 & \#20 & \#20  \\
    \hline
0.98 & 0.80 & 0.85 & 1.00 & 0.95 & 0.99 & 0.91 & 0.87 & 0.75 & 0.80  \\
    \hline
  \end{tabular}}
  \caption{Accuracy of our network on each category in ModelNet40 dataset.}
  \label{Tab10}
\end{table*}

While our network shows its potential in 3D vision, there is still room for further improvement, especially by self-supervised pre-training. The previous works \cite{56,57,60} have validated it is effective to perform self-supervised pre-training on large-scale point cloud datasets for improving the accuracy. However, the compatibility of existing pre-training methods on SSM-based architectures, as well as pre-training techniques tailored for this type of networks remain unexplored. 

In addition, this paper constructs structural dependencies by sorting points along each axis, which destroys geometric structures to a certain extent, and although this problem can be mitigated by merging higher-order interaction features causally inferred from each axis, it may still render the network to struggle to learn geometric structures.

Hence, it is a future endeavor to explore these aspects to make SSM-based works fully outperform attention networks in terms of accuracy in various point cloud tasks.

\begin{table*}[t]
  \centering
  \small
\setlength{\tabcolsep}{7.35mm}{
  \begin{tabular}{cccccccc}
    \hline
Bag & Bin & Box & Cabinet & Chair & Desk & Display & Door  \\
\#100 & \#176 & \#108 & \#360 & \#340 & \#165 & \#174 & \#200  \\
    \hline
0.78 & 0.85 & 0.67 & 0.90 & 0.91 & 0.89 & 0.91 & 0.93  \\
    \hline
Shelf & Table & Bed & Pillow & Sink & Sofa & Toilet &   \\
\#222 & \#233 & \#115 & \#124 & \#115 & \#235 & \#75 &  \\
    \hline
0.90 & 0.86 & 0.86 & 0.87 & 0.91 & 0.92 & 0.88 &   \\
    \hline
  \end{tabular}}
  \caption{Accuracy of our network on each category in ScanObjectNN dataset.}
  \label{Tab11}
\end{table*}

\begin{table*}[t]
  \centering
  \small
\setlength{\tabcolsep}{8.25mm}{
  \begin{tabular}{ccccccc}
    \hline
Ceiling & Floor & Wall & Beam & Column & Window & Door \\
    \hline
96.1 & 98.4 & 87.8 & 0.0 & 44.5 & 65.1 & 75.7  \\
    \hline
Table & Chair & Sofa & Bookcase & Board & Clutter &  \\
    \hline
89.4 & 88.4 & 84.2 & 79.4 & 79.8 & 67.4 &   \\
    \hline
  \end{tabular}}
  \caption{IoU of our network on each category in S3DIS dataset.}
  \label{Tab12}
\end{table*}

\begin{figure*}[!t]
\centering
\includegraphics[width=6.8in]{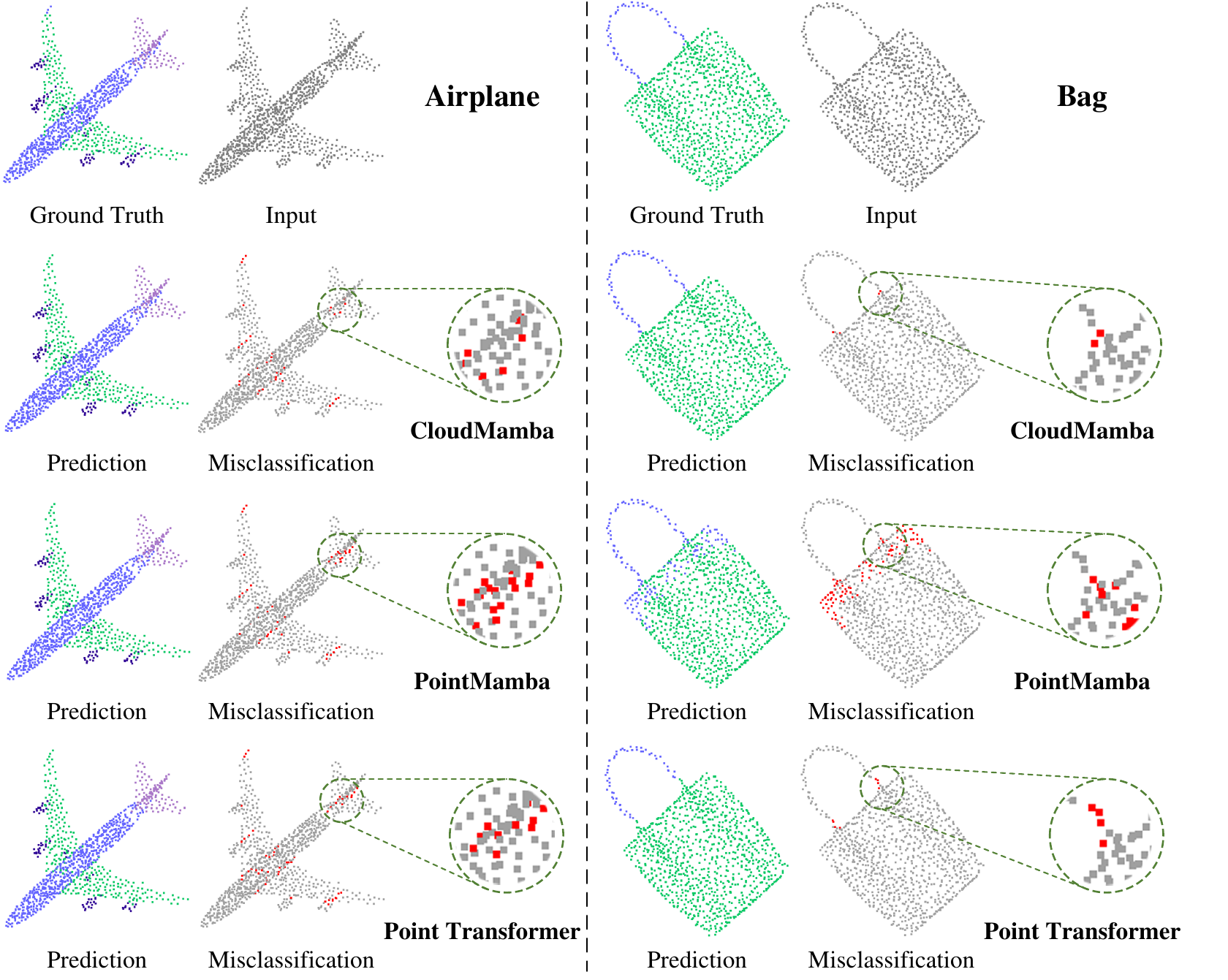}
\end{figure*}

\section*{Appendix I Detailed experimental results}
\label{App9}

For ShapeNet dataset, we list the mIoU of CloudMamba and the comparative networks on each category in Tab. \ref{Tab9} to clearly show the performance of our network on each category. By comparison, we find that CloudMamba achieves the highest mIoU on up to five categories in ShapeNet dataset, and this best result in comparison to the other networks demonstrates our network's ability to effectively leverage and extend Mamba's superior long-range modeling capability, showcasing the potential of SSM in point cloud analysis. In addition, Fig. \ref{Fig6} shows the qualitative results of CloudMamba, PointMamba \cite{38}, and Point Transformer \cite{4} on ShapeNet dataset, which reveals that our network is able to achieve better segmentation results at the boundary of objects. For ModelNet40, ScanObjectNN, and S3DIS datasets, most of the works does not report the accuracy for each category. To clearly demonstrate CloudMamba's performance in each category, as well as to facilitate subsequent studies, Tab. \ref{Tab10}, Tab. \ref{Tab11}, and Tab. \ref{Tab12} list the accuracy of our network for each category on ModelNet40, ScanObjectNN, and S3DIS datasets, respectively. Furthermore, we visualize the segmentation results of CloudMamba for different scenarios in S3DIS dataset, as shown in Fig. \ref{Fig7}.

\begin{figure*}[!t]
\centering
\includegraphics[width=6.8in]{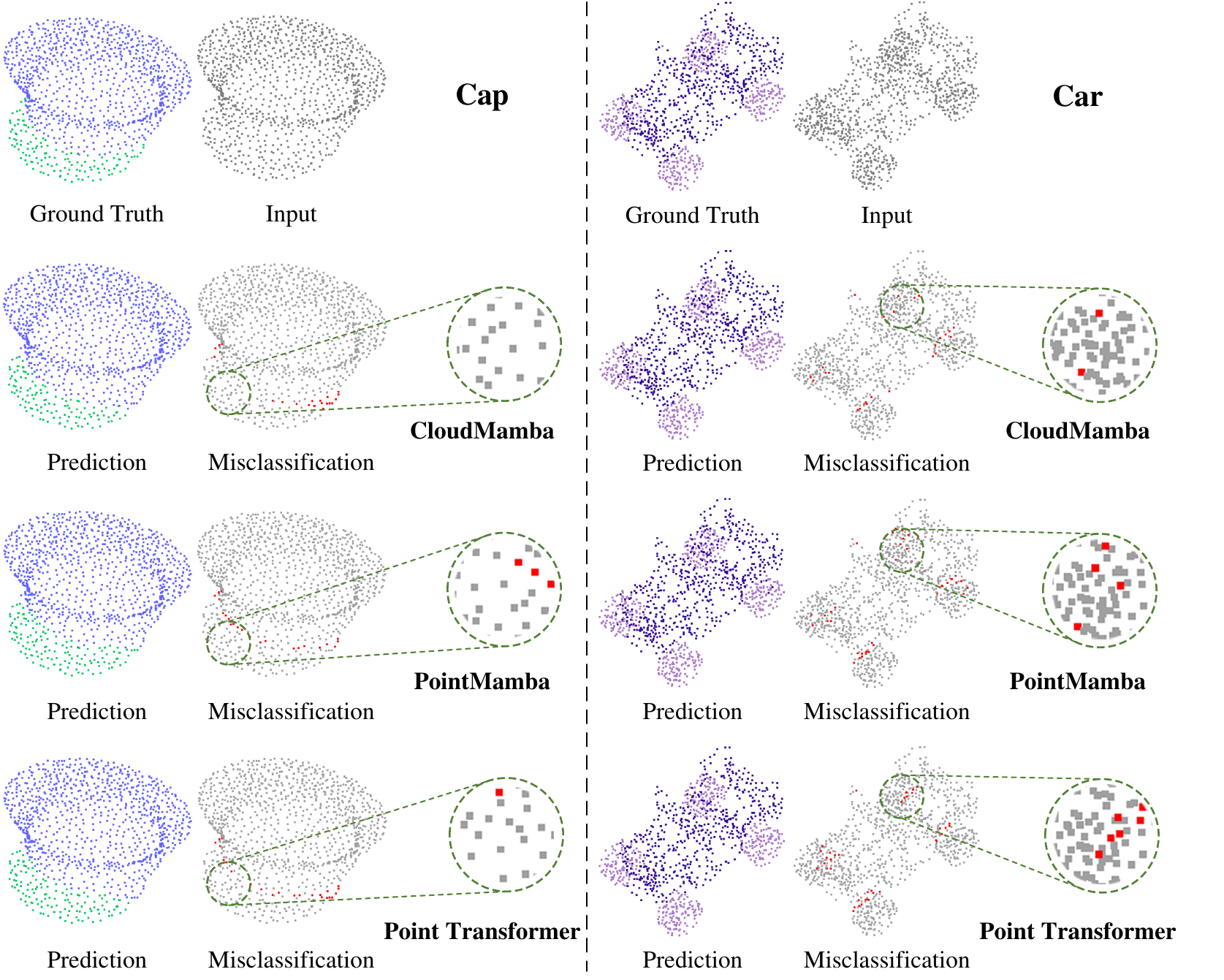}
\end{figure*}

\begin{figure*}[!t]
\centering
\includegraphics[width=6.8in]{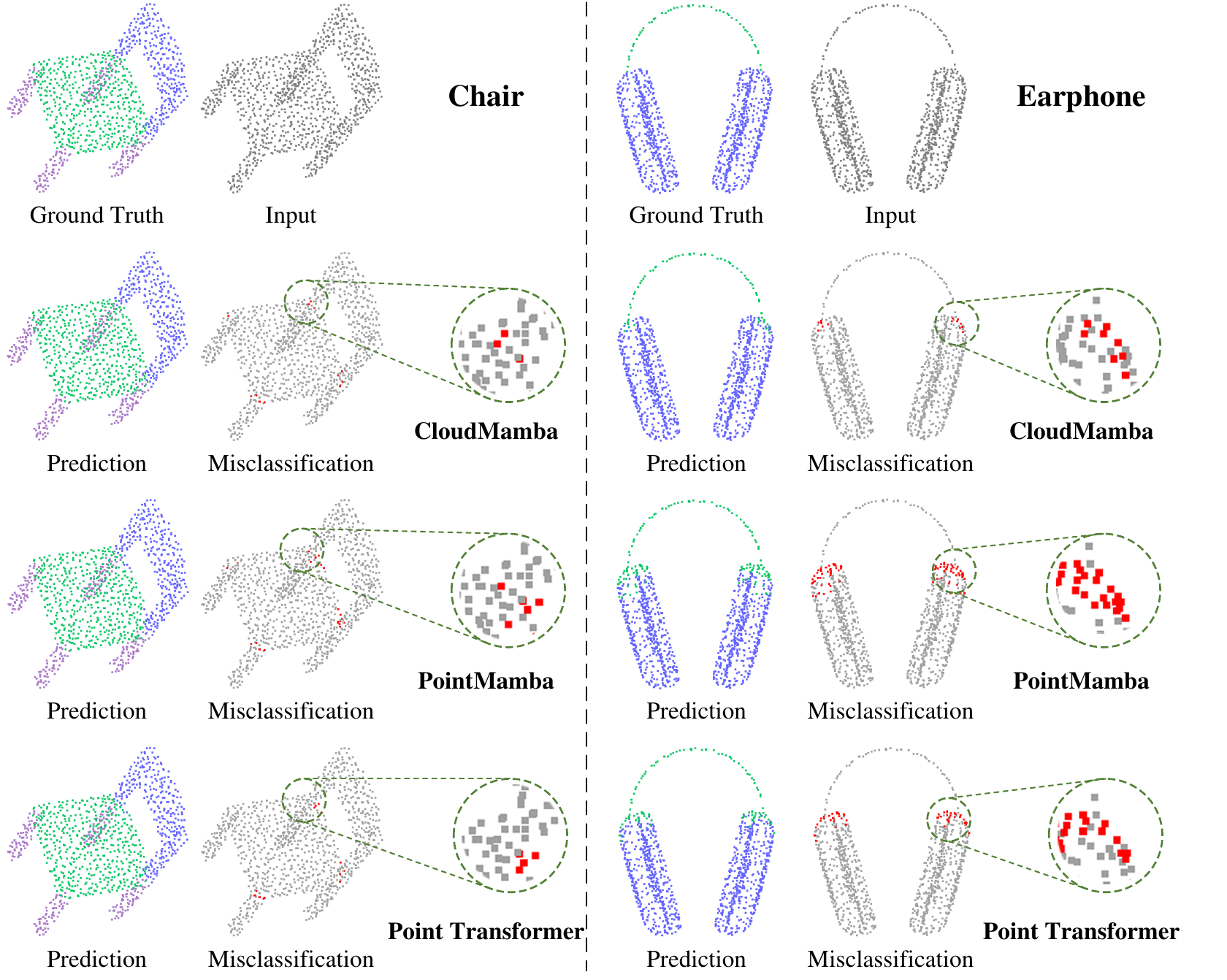}
\end{figure*}

\begin{figure*}[!t]
\centering
\includegraphics[width=6.8in]{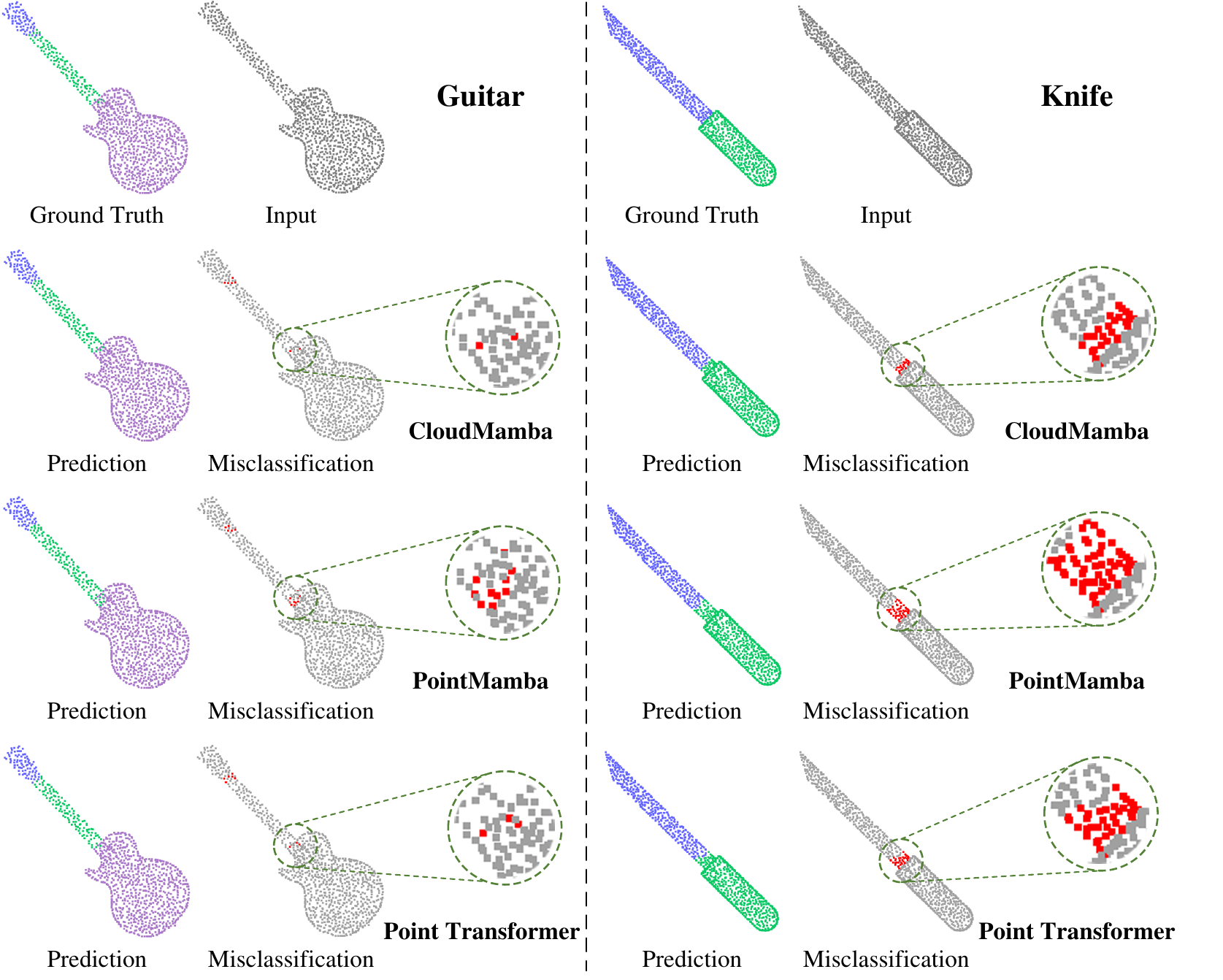}
\end{figure*}

\begin{figure*}[!t]
\centering
\includegraphics[width=6.8in]{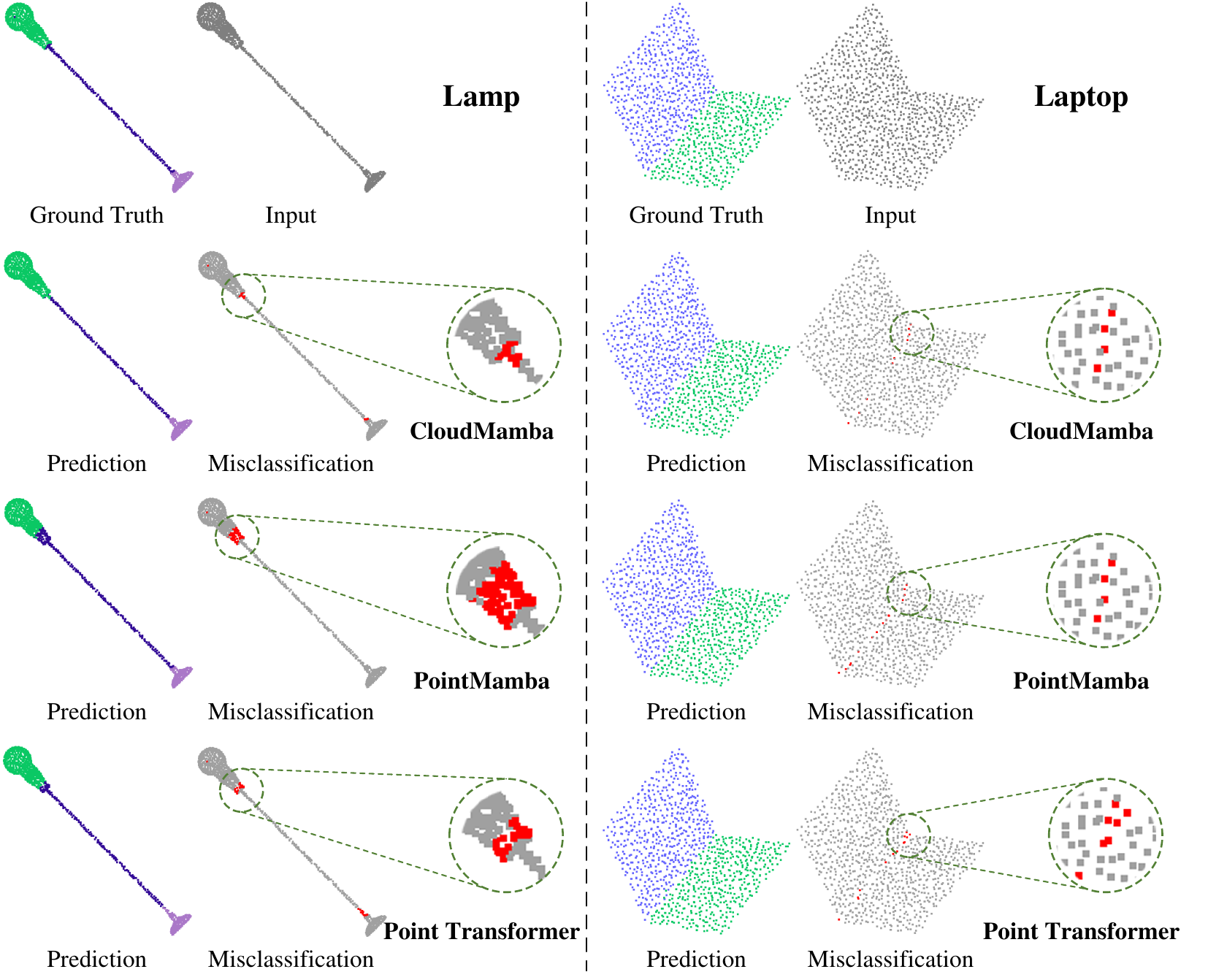}
\end{figure*}

\begin{figure*}[!t]
\centering
\includegraphics[width=6.8in]{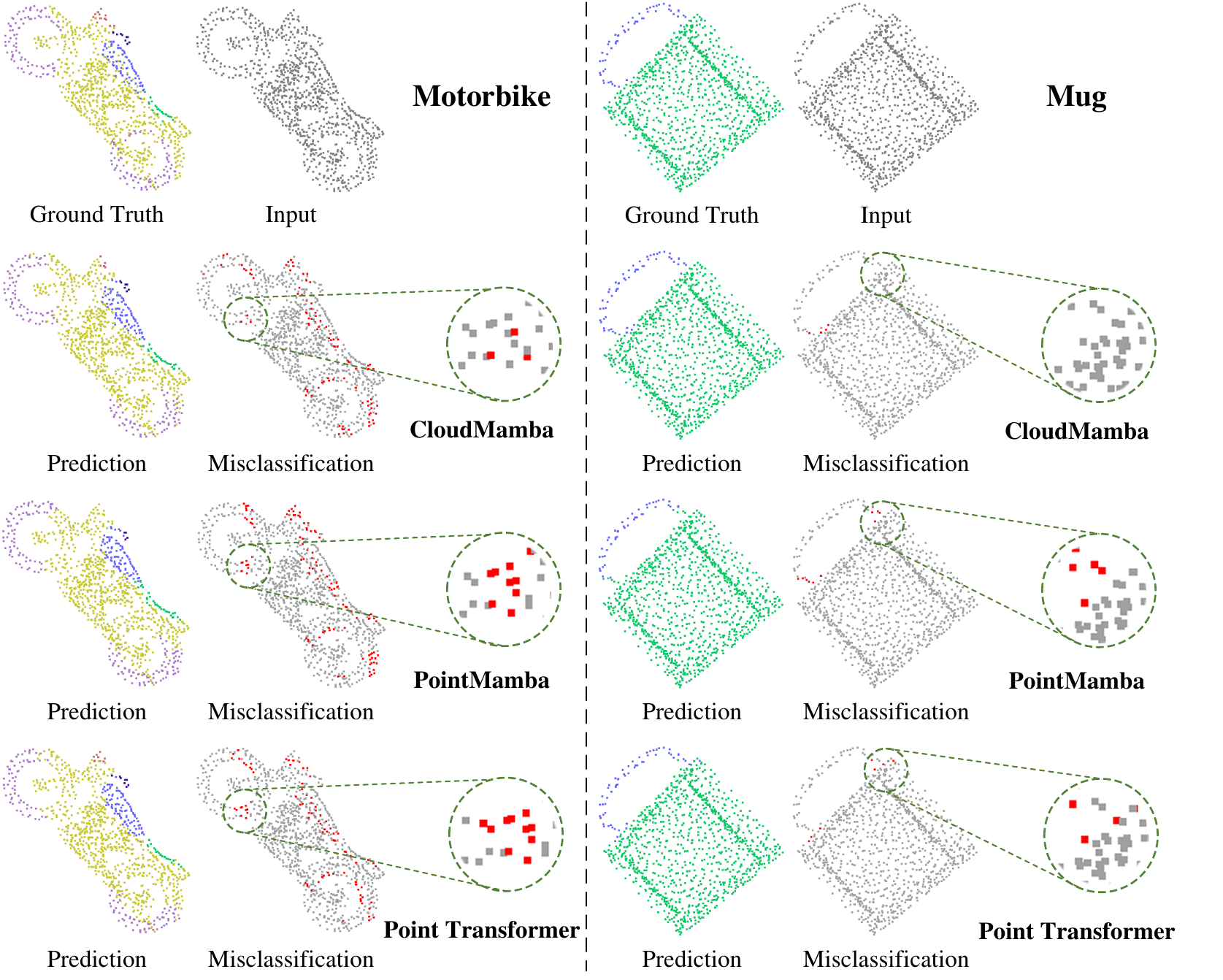}
\end{figure*}

\begin{figure*}[!t]
\centering
\includegraphics[width=6.8in]{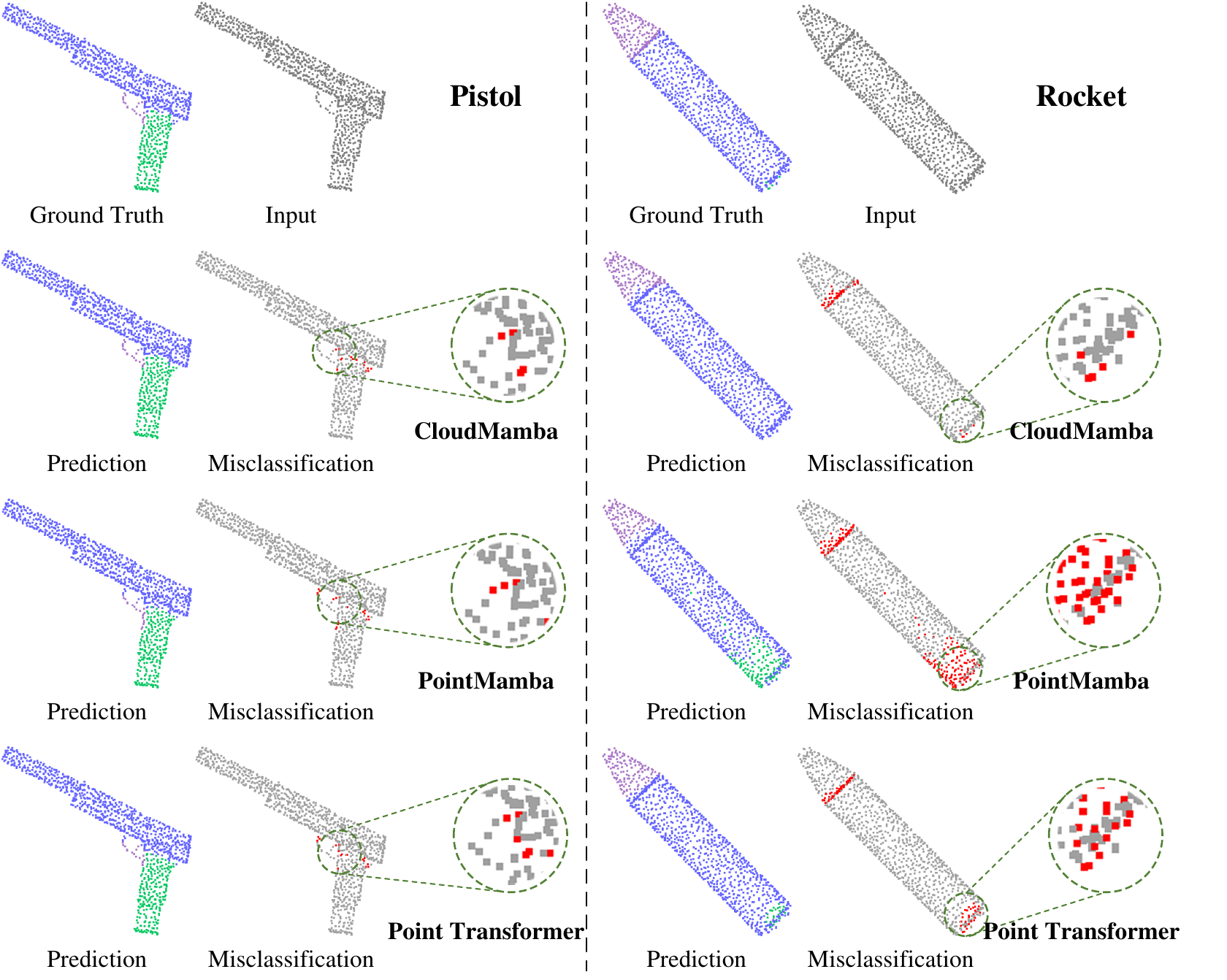}
\end{figure*}

\begin{figure*}[!t]
\centering
\includegraphics[width=6.8in]{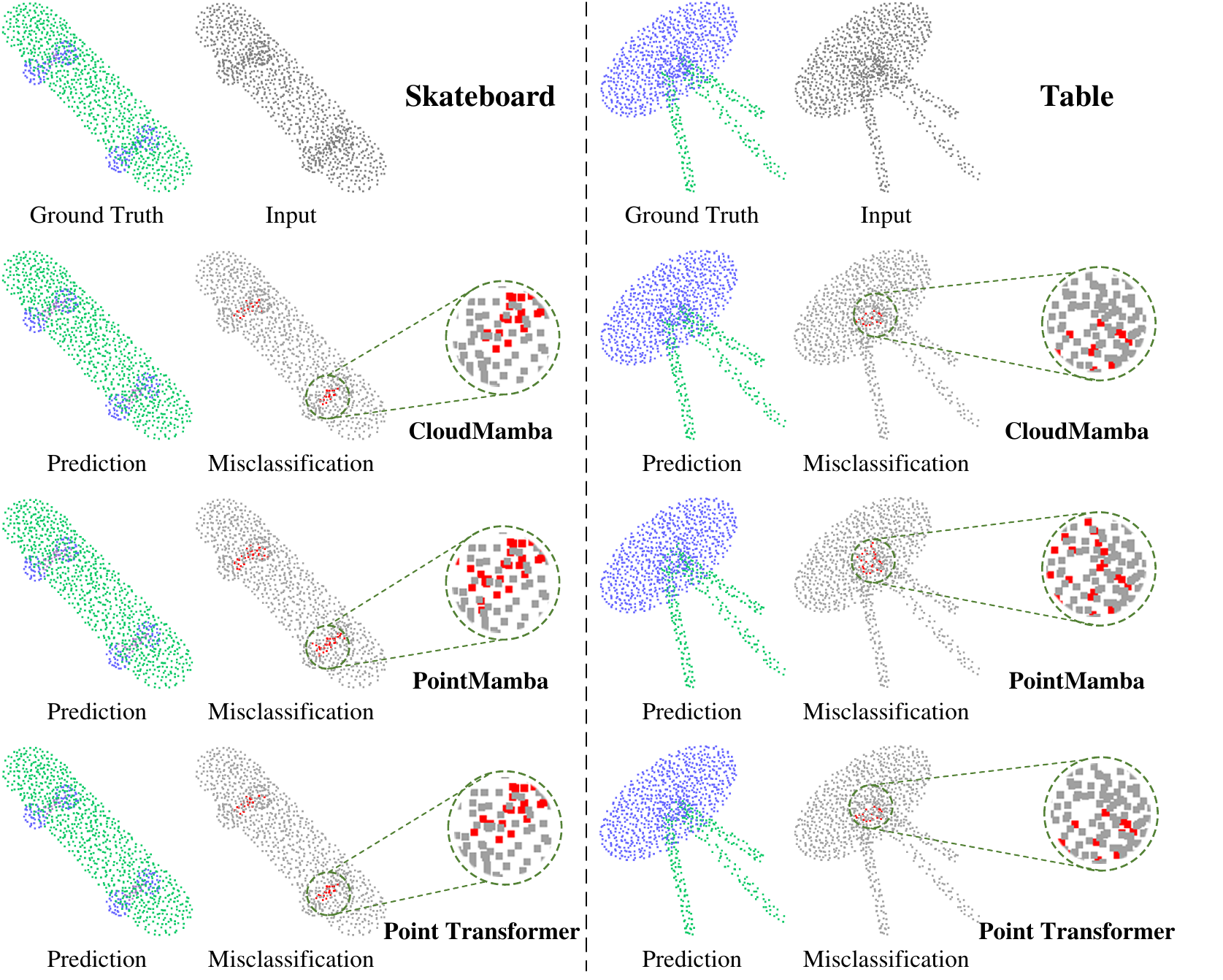}
\caption{Qualitative results of CloudMamba, PointMamba, and Point Transformer on each category in ShapeNet dataset.}
\label{Fig6}
\end{figure*}

\begin{figure*}[!t]
\centering
\includegraphics[height=5.56in]{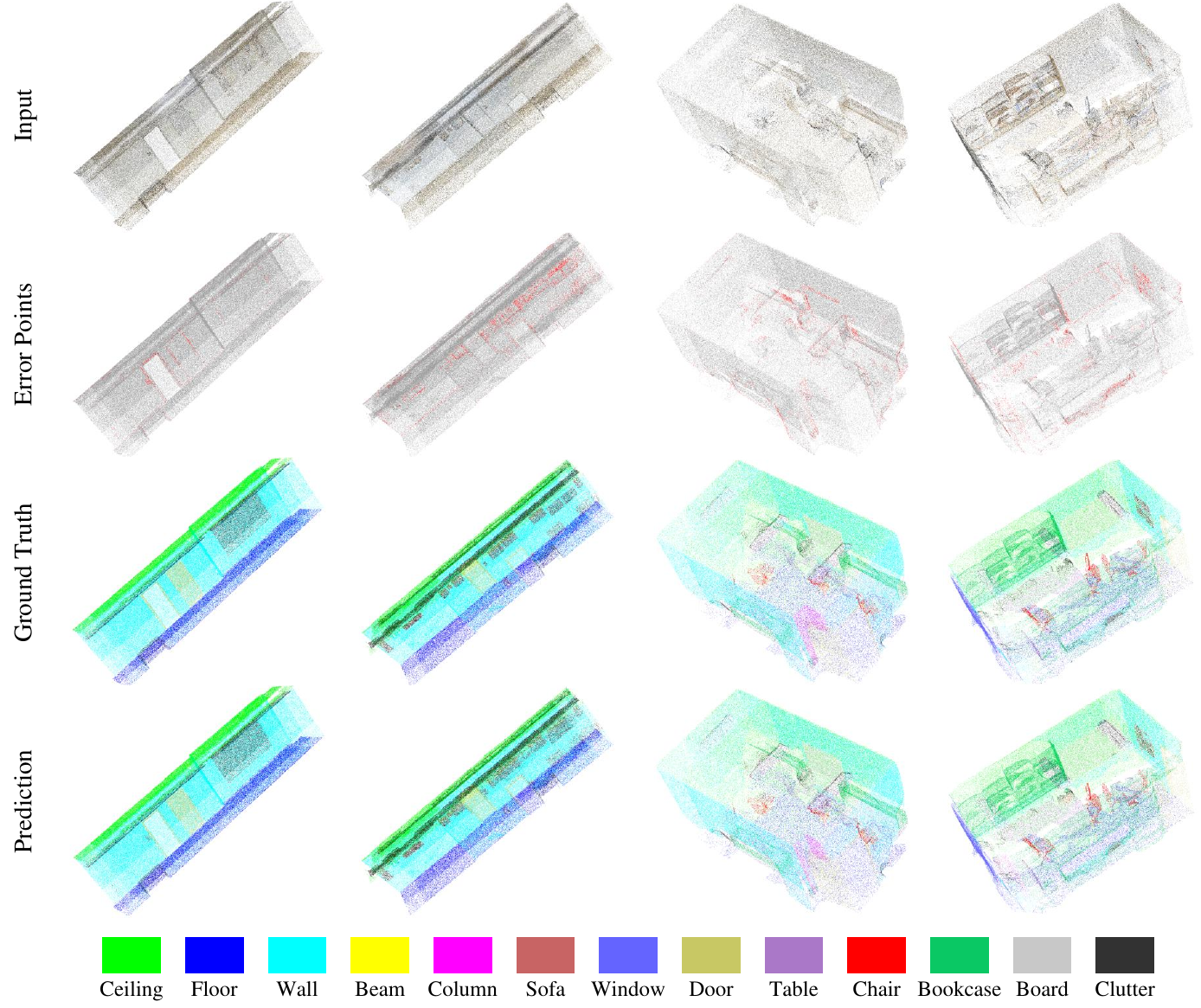}
\end{figure*}

\begin{figure*}[!t]
\centering
\includegraphics[height=5.56in]{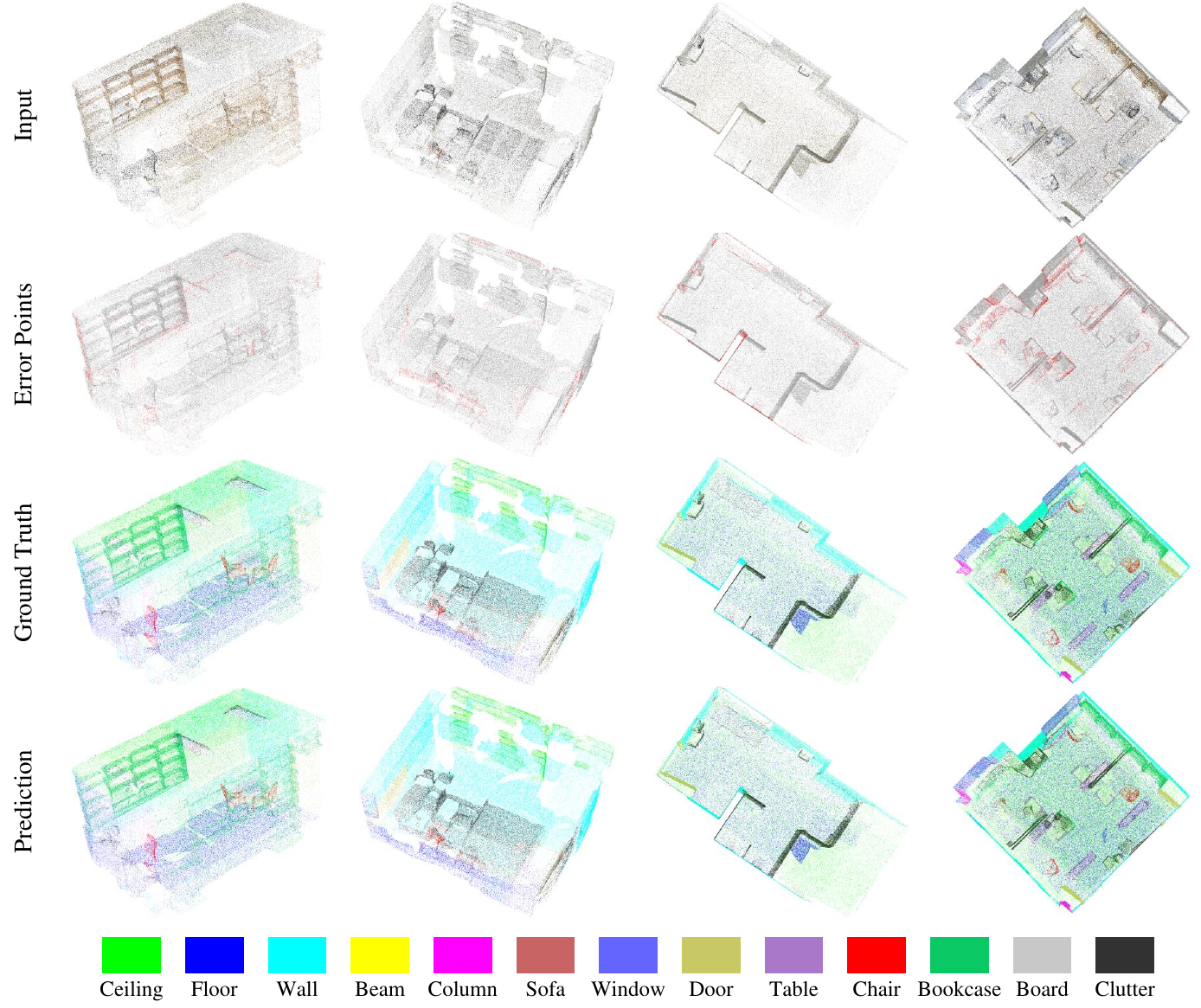}
\caption{Qualitative results of CloudMamba on different scenarios in S3DIS dataset.}
\label{Fig7}
\end{figure*}

\end{document}